\documentclass{article} % For LaTeX2e
\usepackage{iclr2024_conference,times}

\usepackage{amsmath,amsfonts,bm}

\def\1{\bm{1}}

\DeclareMathAlphabet{\mathsfit}{\encodingdefault}{\sfdefault}{m}{sl}
\SetMathAlphabet{\mathsfit}{bold}{\encodingdefault}{\sfdefault}{bx}{n}

\newcommand{\softmax}{\mathrm{softmax}}

\usepackage{hyperref}
\usepackage{url}

\usepackage{microtype}
\usepackage{enumitem}
\usepackage{graphicx}
\usepackage{float}
\usepackage{mathrsfs}
\usepackage{color}
\usepackage{csquotes}
\usepackage{spverbatim}
\UseRawInputEncoding
\usepackage{verbatim}
\usepackage{booktabs}
\usepackage{multirow}
\usepackage{algorithm}
\usepackage{algorithmic}
\usepackage{caption}
\usepackage{subcaption}
\usepackage{MnSymbol}
\usepackage{pifont}
\usepackage{natbib}
\usepackage{subcaption,siunitx,booktabs}
\usepackage{cleveref}
\setcitestyle{nameyear,citesep={;},aysep={,},yysep={;}}
\newcommand{\xmark}{\small\text{\ding{54}}}
\newcommand{\reddiamond}{{\color{red}\ding{169}}}

\definecolor{sky}{RGB}{0, 230, 230}
\definecolor{violetcolor}{RGB}{239, 66, 245}
\definecolor{greencolor}{RGB}{50,205,50}
\definecolor{darkgreencolor}{RGB}{0,128,0}
\definecolor{orangecolor}{RGB}{242, 136, 34}
\definecolor{redcolor}{RGB}{215,25,28}

\usepackage{xspace}

\makeatletter
\DeclareRobustCommand\onedot{\futurelet\@let@token\@onedot}
\def\@onedot{\ifx\@let@token.\else.\null\fi\xspace}
\def\eg{\emph{e.g}\onedot} 
\def\ie{\emph{i.e}\onedot}

\makeatother

\makeatletter
\newenvironment{myprompt}[1]
 {\def\@xobeysp{\ }\verbatim\rightskip=0pt plus 1em\relax
   \small{  \input{#1} } }
 {\endverbatim}
\makeatother

\ifx\BlackBox\undefined
\newcommand{\BlackBox}{\rule{1.5ex}{1.5ex}}  % end of proof
\fi

\ifx\QED\undefined
\def\QED{~\rule[-1pt]{5pt}{5pt}\par\medskip}
\fi

\ifx\proof\undefined
\newenvironment{proof}{\par\noindent{\bf Proof\ }}{\hfill\BlackBox\\[2mm]}
\fi

\ifx\theorem\undefined
\newtheorem{theorem}{Theorem}
\fi
\ifx\example\undefined
\newtheorem{example}[theorem]{Example}
\fi
\ifx\property\undefined
\newtheorem{property}{Property}
\fi
\ifx\lemma\undefined
\newtheorem{lemma}[theorem]{Lemma}
\fi

\def\##1\#{\begin{align}#1\end{align}}
\def\$#1\${\begin{align*}#1\end{align*}}

\def\emb{\mathrm{emb}}
\def\vocab{\mathtt{vocab}}
\def\prompt{\mathtt{prompt}}
\def\response{\mathtt{res}}
\def\cipher{\mathtt{cipher}}
\def\logit{\mathrm{logit}}
\def\softmax{\mathrm{softmax}}

\newcommand{\cV}{\mathcal{V}}

\newcommand{\RR}{\mathbb{R}}

\newcommand*\samethanks[1][\value{footnote}]{\footnotemark[#1]}
\newcommand{\Hquad}{\hspace{1em}}

\title{Let Models Speak Ciphers:\\ Multiagent Debate through Embeddings}

\author{%
  \normalfont Chau Pham$^1$\thanks{Equal contribution; work done during an internship at ByteDance.} \qquad Boyi Liu$^2$\samethanks[1] \qquad Yingxiang Yang$^3$ \qquad Zhengyu Chen$^3$ \qquad Tianyi Liu$^3$\\ \qquad Jianbo Yuan$^3$ \qquad Bryan A. Plummer$^1$\thanks{Equal advising.} \qquad Zhaoran Wang$^2$\samethanks[2] \qquad Hongxia Yang$^3$\samethanks[2]\\
  $^1$Boston University, \Hquad $^2$Northwestern University,  \Hquad  $^3$ByteDance Inc.\\
 \footnotesize \texttt{\{chaupham,bplum\}}@bu.edu\\
 \footnotesize \texttt{boyiliu2018}@u.northwestern.edu, \texttt{zhaoranwang}@northwestern.edu \\
 \footnotesize \texttt{\{yingxiang.yang,zhengyu.chen,tianyi.liu,jianbo.yuan,hx.yang\}}@bytedance.com
}

\iclrfinalcopy % Uncomment for camera-ready version, but NOT for submission.
\begin{document}

\maketitle

\begin{abstract}
Discussion and debate among Large Language Models (LLMs) have gained considerable attention due to their potential to enhance the reasoning ability of LLMs. Although natural language is an obvious choice for communication due to LLM's language understanding capability, the token sampling step needed when generating natural language poses a potential risk of information loss, as it uses only one token to represent the model's belief across the entire vocabulary. In this paper, we introduce a communication regime named CIPHER (\underline{C}ommunicative \underline{I}nter-Model \underline{P}rotocol T\underline{h}rough \underline{E}mbedding \underline{R}epresentation) to address this issue. Specifically, we remove the token sampling step from LLMs and let them communicate their beliefs across the vocabulary through the expectation of the raw transformer output embeddings. Remarkably, by deviating from natural language, CIPHER offers an advantage of encoding a broader spectrum of information without any modification to the model weights, outperforming the state-of-the-art LLM debate methods using natural language by $0.5-5.0$\% across five reasoning tasks and multiple open-source LLMs of varying sizes. This showcases the superiority and robustness of embeddings as an alternative ``language" for communication among LLMs. We anticipate that CIPHER will inspire further exploration for the design of interactions within LLM agent systems, offering a new direction that could significantly influence future developments in the field.
\end{abstract}

\section{Introduction}
\label{sec:intro}

%%% Task & SOTA
Recent studies in Large Language Models (LLMs) have demonstrated tremendous potential of LLMs to enhance the quality of their responses on reasoning tasks through discussion and debate \citep{chen2023teaching,madaan2023selfrefine,paul2023refiner,fu2023improving,jiang2023selfevolve,du2023improving,liang2023encouraging}. However, these approaches are often only effective for state-of-the-art LLMs such as GPT-4 \citep{OpenAI2023}, and have not proven successful with smaller and open-source models such as Vicuna-13B \citep{vicuna2023}. For example, \cite{olausson2023demystifying} found that using self-repair in code generation tasks, is only effective with GPT-4, but proves ineffective when applied to GPT-3.5~\citep{gpt35}. Similarly, \cite{fu2023improving} tested the ability of LLMs in a bargaining game setup and discovered that only a few well-aligned models such as GPT-4 and Claude-v1.3 \citep{claude} can continuously improve their responses by utilizing feedback. Such inconsistencies in performance across various LLMs motivate our pursuit of a universal solution with consistent efficacy irrespective of the specific LLM employed.

\begin{figure}[t]
\vskip-0.45in
\begin{center}
\centerline{\includegraphics[width=0.85\columnwidth]{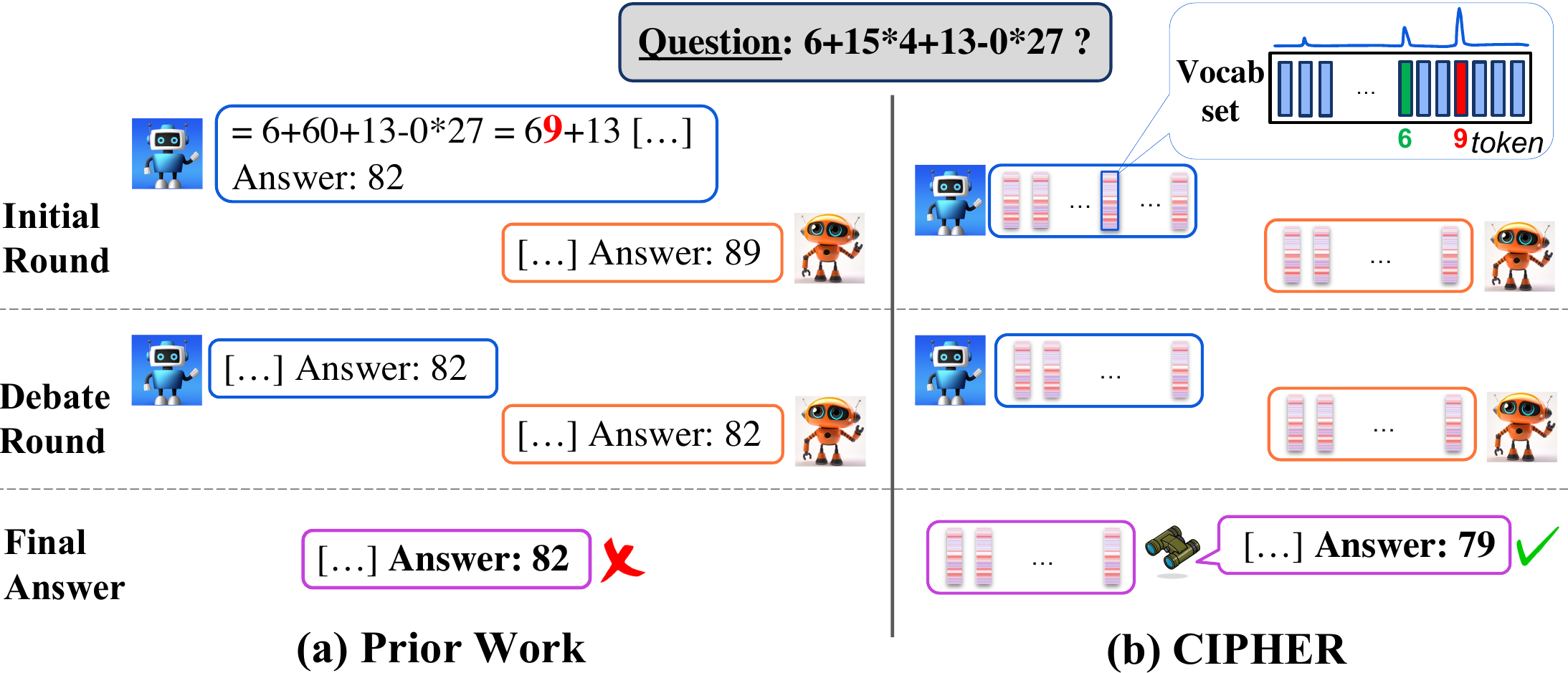}}
% \vskip-0.1in
\caption{\textbf{Comparison of different communication regimes.} \textbf{(a)} LLaMA-2-70B \citep{touvron2023llama} makes a mistake by generating token ``\textbf{\textcolor{red}{9}}", which should be ``\textbf{\textcolor{darkgreencolor}{6}}" instead. This shows natural language communication between LLMs can lose information. \textbf{(b)} In CIPHER (ours), the model outputs embedding vectors and directly receives vectors from other models as inputs. 
Specifically, instead of sampling one token as seen in prior work, CIPHER generates a vector by taking weighted average of all tokens' embeddings in the vocabulary set. Our vectors provide a richer source of information while also being human interpretable by mapping them back to natural language via a nearest neighbor search over the vocabulary.}
\label{fig:debate_demo}
\end{center}
\vskip-0.2in
\end{figure}

To leverage LLM's language understanding capability, prior work~\citep{du2023improving,liang2023encouraging} adopted natural language during debates. 
Using natural language for debate is appealing due to its interpretability, but we argue that natural language is neither necessary nor optimal for inter-LLM communication. First, we do not necessarily need to understand the intermediate debates amongst LLMs. Second, natural language generation uses only one token to represent the model's belief over the entire vocabulary, which risks losing information embedded within the model output logits. For instance, in reasoning tasks, the most confident token can be wrong. As shown in the top row of Fig.~\ref{fig:debate_demo}a, the model makes a mistake in generating the token ``\textbf{\textcolor{red}{9}}" by picking the most confident token while discarding the valuable information contained in the correct token ``\textbf{\textcolor{darkgreencolor}{6}}".

We address this by proposing a novel communication protocol, \underline{C}ommunicative \underline{I}nter-Model \underline{P}rotocol T\underline{h}rough \underline{E}mbedding \underline{R}epresentation (CIPHER), that enables LLMs to communicate more freely in the tokenizer's embedding space in the multiagent debate setting. To tackle the issue of information loss in natural language debates, CIPHER lets the models communicate in a vector space, which represents the entire space of potential outputs, thus resulting in a further boost in the performance of debate (Fig.~\ref{fig:debate_demo}b). Specifically, it bypasses the token sampling process by generating the weighted average of all tokens' embeddings in the vocabulary set, instead of yielding a single token. Thus, this passes richer information to other models during debate, particularly when the model is uncertain. While this form of communication deviates from natural language, it still remains interpretable to humans via nearest neighbor search over the vocabulary.

During a debate, LLMs begin by independently providing initial responses to the question (\textit{Initial Round}). Then, each agent receives the other agent's answer to refine its previous answer (\textit{Debate Round}). After some debate rounds, the agents ultimately reach a consensus on a final answer in the majority of cases \citep{du2023improving}.
We show that LLMs can effectively communicate without relying on natural language and even achieve superior performance.

We validate this claim by evaluating our approach on five diverse datasets across multiple domains: GSM8K~\citep{cobbe2021training}, Arithmetic \citep{du2023improving}, MMLU Formal Logic, MMLU High School Math, and MMLU Professional Psychology \citep{hendrycks2020measuring}. Our experiment results show {$0.5-5.0\%$} performance improvement over the natural-language counterpart. The results align with prior work indicating that when communicating through natural language, weaker models do not outperform majority voting approaches through debates or self-refinement~\citep{wang2022self}. Criticality, in contrast to prior work \citep{madaan2023selfrefine,du2023improving,liang2023encouraging}, we find that our approach can generalize across a wide array of LLMs, enabling even smaller LLMs to unlock the benefits of debate and achieve better performance than majority voting~\citep{wang2022self}. This suggests that, even for open-source models, debate is still an efficient form of communication to boost the performance of LLMs. In summary, our contributions are three-fold:

\begin{itemize}[nosep,leftmargin=*]
    \item We propose CIPHER, a novel inter-model communication protocol for LLMs that share the same tokenizer, regardless of whether they use identical or different embedding-vocabulary mappings.
    \item We perform comprehensive experiments to validate the efficacy of state-of-the-art debate method through natural language communication \citep{du2023improving} and CIPHER. Our results show that even less powerful LLMs can still benefit from debates.
    \item We conduct an extensive ablation study to shed light on the mechanisms that make communication through embeddings more effective for debates among LLMs.
\end{itemize}

\section{Related Work}
\label{sec:related_work}

\noindent\textbf{Multiagent debate.} Multiagent debates was pioneered by \citet{du2023improving}, where LLMs provide initial responses and then make refinements by iteratively considering inputs from peers. Typically, the LLMs reach a consensus in a few debate rounds that is often more accurate. In a concurrent study, \citet{liang2023encouraging} incorporated a ``judge" to resolve tiebreakers and determine the final answer. While closely related to the self-improvement via feedback approach where a model iteratively refines its own responses (\eg, ~\citet{madaan2023selfrefine,Akyurek2023RL4FGN}), this debate approach involves multiple agents with the same role. Thus, it not only encourages divergent thinking in LLMs, but also eliminates the bottleneck at the feedback step, resulting in superior performance compared to self-improvement via feedback methods across diverse datasets. Despite these advances, prior work on multiagent debate  have only focused on large and closed-source models such as GPT-4 and GPT-3.5, leaving the efficacy of debate on smaller, open-source models underexplored. To address this gap, our study adapts a multiagent debate setup of \citet{du2023improving} and introduce a novel communication protocol in which agents can interact without using natural language.

\noindent\textbf{Self-improvement via feedback.} \cite{madaan2023selfrefine} presented a self-improvement framework, where an LLM doubles as both generator and critic. While this approach allows models such as GPT-3.5 and GPT-4 to boost performance, it falters for smaller and less competent models such as Vicuna-13B \citep{vicuna2023}, which struggled on consistently generating feedback in the required format. Concurrently, \citet{Akyurek2023RL4FGN} introduced RL4F, sidestepping generator weight updates, while \citet{shinn2023reflexion} exploited verbal reinforcement for error correction. \citet{fu2023improving} applied self-improvement in a bargaining context, noting a distinct advantage for models like GPT-4 and Claude-v1.3 in iterative improvement. Overall, this body of work highlights the necessity for powerful LLMs as critics and the prevailing limitations in models with parameters fewer than 52B in incorporating natural language feedback effectively \citep{bai2022constitutional,saunders2022self}. 

\noindent{\bf Self-debugging for code generation.} \citet{chen2023teaching, jiang2023selfevolve,olausson2023demystifying} focused on improving coding ability of LLMs by letting them incorporate explanation of the generated code in natural language and execution results from unit tests to self-debug. Similar to the aforementioned studies, they revealed that models that are more powerful than the generator model yields better results when used as the critic (e.g., GPT-4 gives feedback to GPT-3.5, or humans give feedback to GPT-4). Additionally, \citet{olausson2023demystifying} reported that self-repair on weaker models cannot improve over majority voting \citep{wang2022self}. In short, these studies underscore a bottleneck in the critic role, necessitating the use of powerful LLMs for generating valuable feedback. 

\noindent\textbf{Reasoning ability in Language Models via prompting.} To further strengthen the reasoning ability of LLMs, \citet{wei2022chain} proposed Chain-of-Thought (CoT), a method that employs a series of intermediate reasoning steps to incrementally achieve the final goal. This idea was subsequently generalized by \citet{yao2023tree,long2023large} into Tree-of-Thought, which explores multiple different intermediate steps for the best reasoning path and backtracks when necessary. Graph-of-Thought \citep{besta2023graph} further extends the reasoning ability of LLMs by considering the LLM reasoning as an arbitrary graph, where vertices and edges represent thoughts and their dependencies, respectively.
CIPHER incorporates both CoT and few-shot CoT into prompt engineering for boost performance.

\section{CIPHER: Communicative Inter-Model Protocol through Embedding Representation
}\label{sec:propsed_method}
LLMs function by taking a prompt as input and autoregressively generating a sequence of tokens as the response. From a tokenizer, we have a vocabulary set $\cV = \{\vocab_i\}_{i \in [V]}$ where $[V] = \{1, \dots, V\}$ is an index set. Let $\prompt$ be the prompt, which contains the instruction, question, and (possible) responses collected in previous debate rounds. We define $\response$ to be the generated response from the LLMs. We use $(t)$ in superscripts to indicate the step $t$ of response generation. Correspondingly, we use $(1:t)$ in superscripts to indicate the concatenation of the tokens (or embeddings) generated during the first $t$ steps. A variable is empty whenever it has superscript $(1:t)$ with $t < 1$. 

\subsection{Natural Language Communication}\label{sec:nl_bad}
When generating responses, causal LLMs (e.g., LLaMA~\citep{touvron2023llama1}) 
generate tokens one at a time based on the words prior to each token. Given $\prompt$ and the first $t-1$ generated response tokens $\response^{(1:t-1)}$, the causal LM calculates $\logit(\prompt, \response^{(1:t-1)} ) \in \RR^{1 \times V}$, which is a vector of logits. Then, the next token is sampled from the vocabulary $\cV$ with respect to the distribution

\vskip -0.15in
\#\label{eq:genprob}
p^{(t)} = [p_{\vocab_1}^{(t)}, \dots, p_{\vocab_V}^{(t)} ] = \softmax \bigl\{\logit(\prompt, \response^{(1:t-1)})/T \bigr\},
\#
where $T > 0$ is the temperature. The distribution $p^{(t)}$ can be viewed as the LLM's belief regarding the most appropriate token at the current position. However, the token sampling step compresses the information of $p^{(t)}$ into a single token, discarding the information on all other tokens in the process. While using one token at each position is useful for humans to understand the outputs from LLMs, we posit that it is not a requirement for effective inter-LLM communication. Recall that in Fig.~\ref{fig:debate_demo}b, we observe the rich information contained in the LLM's belief by the probability distribution over all tokens. Thus, we argue that the token sampling process, which sacrifices information for readability, may lead to a sub-optimal solution for inter-LLM communication. In light of this consideration, we present our CIPHER response generation method for multiagent debate in the section below.

\subsection{Communication through Semantic Embeddings}\label{sec:cipher}

\begin{figure}[tb]
\vskip -0.45in
\begin{center}
\centerline{\includegraphics[width=0.85\columnwidth]{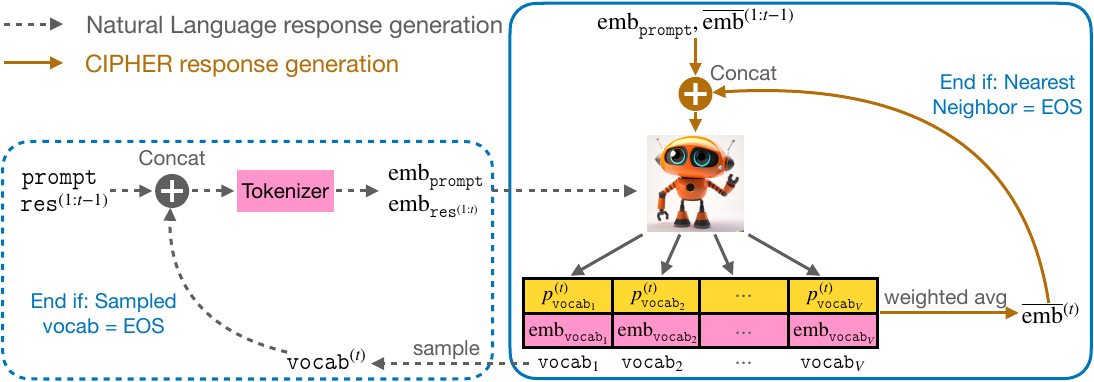}}
\vspace{-2mm}
\caption{\textbf{Response generation in Natural Language and CIPHER}. {\it Natural Language response generation:} Starting from the top left of the figure, the tokenizer encodes the texts of $\prompt$ and $\response^{(1:t-1)}$ into embedding inputs for the LLM. The LLM then outputs a distribution $p^{(t)} = [p_{\vocab_1}^{(t)}, \dots, p_{\vocab_V}^{(t)}]$ over $\cV$, from which the next token $\vocab^{(t)}$ is sampled (bottom mid). {\it CIPHER response generation:} Instead of sampling a single token $\vocab^{(t)}$ at step $t$, CIPHER generates a weighted average embedding vector $\overline{\emb}^{(t)}$, utilizing $p^{(t)}$ as weights (Eq.~\ref{eq:emb-response}). The embedding vector, together with previously generated embeddings $\overline{\emb}^{(1:t-1)}$ and $\prompt$ embedding $\emb_{\prompt}$, bypasses the tokenizer decoding step and circles directly back into the LLM. Note that CIPHER-generated semantic embeddings approximate a token embedding, but do not precisely match it.}
\label{fig:cipher}
\end{center}
\vskip -0.2in
\end{figure}

We propose CIPHER, an embedding communication protocol for LLM debates to capture richer encoded information. Our goal is to encode as much information as possible during inter-LLM communication. However, LLMs are designed to understand natural language sequences. Thus, they might not be able to grasp vectors that reside outside the convex hull of the tokenizer's embedding space. To address this, we propose to use the weighted average of embeddings in place of the tokens sampled with respect to $p^{(t)}$ (Eq.~\ref{eq:genprob}) in the autoregressive response generation process (Fig.~\ref{fig:cipher}).

\noindent{\bf CIPHER response generation.} Let $\emb_{\prompt}$ be the embeddings of $\prompt$, and $\emb_{\vocab_i}  \in \RR^{1 \times d}$ be the embedding of token $\vocab_i \in \cV$. Given the concatenation $\overline{\emb}^{(1:t-1)}$ of previously $t-1$ generated semantic embeddings, CIPHER recursively generates new semantic embedding at each step $t = 1, 2, \dots$ by
\vskip -0.4in
\#
\overline{\emb}^{(t)} & = \sum_{i = 1}^{V} \quad p^{(t)}_{\vocab_i} \cdot \emb_{\vocab_i},\label{eq:emb-response}\\
\text{where} \quad [p^{(t)}_{\vocab_1}, \dots, p^{(t)}_{\vocab_V}] & = \softmax \big\{\logit(\emb_\prompt, \overline{\emb}^{(1:t-1)})/T \bigr\}.\label{eq:cipher_prob}
\#
Here $T>0$ is the temperature. While the probability $p^{(t)}_{\vocab_i}$ in Eq.~\ref{eq:cipher_prob} bears resemblance to that in Eq.~\ref{eq:genprob}, CIPHER bypasses the tokenizer and directly passes semantic embeddings into LLMs instead of taking raw texts as inputs. The generation process stops either (i) the EOS token embedding becomes the nearest neighbor of the newly generated embedding, or (ii) the maximal sequence length is reached. For a CIPHER response generation process that stops after $\tau$ steps, the CIPHER response is defined as $\cipher = \overline{\emb}^{(1:\tau)}$, which is the concatenation of all the generated semantic embeddings.

\noindent{\bf CIPHER debate.} We formalize the CIPHER debate procedure in Algorithm \ref{alg_cipher}. First, we convert the question and instructions into embeddings $\emb_{\prompt}$ using the tokenizer (Line 2). Then, for each debate round, we form an embedding representation by concatenating $\emb_{\prompt}$ and (possible) CIPHER responses, $\cipher_{i}$, from all debaters in previous rounds (Line 4). This embedding representation is then input directly into the models without the token decoding step. The debaters then generate refined CIPHER responses following Eq.~\ref{eq:emb-response} (Line 5). To close the debate, at \textit{Convert-and-Aggregate} step (Line 6), we convert the embedding responses back to natural language using nearest neighbor search over the vocabulary set, then aggregate them to obtain the final response. In most cases, LLM debaters typically reach a consensus answer by the final round, as observed in~\citet{du2023improving}. When divergence in final responses occurs, majority voting \citep{wang2022self} or random tie-breaking are often used. However, majority voting may not be suitable for open-ended questions (\eg, summarization tasks) where multiple correct answers exist, as in \textit{the game of 24} \citep{yao2023tree}, and scenarios where debates involving only two agents. Thus, in our experiments, we select the response from the debater with the lowest temperature as the final answer. This approach achieves comparable accuracy to the best performing debater while maintaining computational efficiency by running inference on only one model in the final round, as shown in Fig.~\ref{fig:contour_plots}.

\begin{algorithm}[t]\algsetup{linenosize=\tiny}\small
\caption{CIPHER Debate}
\label{alg_cipher}
\begin{algorithmic}[1]
{\small
\STATE {\bf Input:}
Question and instructions $\prompt$, number of rounds $R \geq 2$, and $n$ CIPHER debaters $\{D_i\}_{i \in [n]}$.
\\-------------------------------------------------------------- (\textit{initial round}) -------------------------------------------------\vspace{-1pt}

{\STATE
Obtain embedding of the prompt, $\emb_{\prompt}$, via the tokenizer
}

{\bf For debater $i=1,\dots,n$:}

{\addtolength{\leftskip}{0.2in}

\STATE Get initial CIPHER response $\cipher_i \leftarrow D_i(\emb_{\prompt})$ from debater $i$ by \eqref{eq:emb-response}

}

{\bf EndFor}
\\------------------------------------------------------------- (\textit{debate rounds}) -----------------------------------------------\vspace{-1.5pt}

% Debate rounds
{\bf For round $r=2,\dots,R$:}

{\addtolength{\leftskip}{0.2in}

\STATE Get updated prompt embedding $\emb_\prompt \leftarrow \mathrm{concat}(\emb_{\prompt}, \cipher_1, \dots, \cipher_n)$

}

{\addtolength{\leftskip}{0.2in}

{\bf For debater $i=1,\dots,n$:}

}

{\addtolength{\leftskip}{0.4in}

\STATE Get CIPHER response $\cipher_i \leftarrow D_i(\emb_{\prompt})$ from debater $i$ by \eqref{eq:emb-response}

}
% {\bf EndFor}

% }
{\addtolength{\leftskip}{0.2in}

{\bf EndFor}

}
{\bf EndFor}
\\------------------------------------------------------------ (\textit{post processing}) ----------------------------------------------\vspace{-1pt}

\STATE {\bf Output:} Final response $\cipher^* = \textit{\text{Convert-and-Aggregate}}(\cipher_1 , \dots , \cipher_n)$
}

\end{algorithmic}
\end{algorithm}

\noindent{\bf Role of temperature.} The temperature $T$ in Eq.~\ref{eq:genprob} and Eq.~\ref{eq:cipher_prob} controls the smoothness of the probability $p^{(t)}_{\vocab_i}$. When $T \rightarrow 0$,  both CIPHER's embedding generation and natural language generation result in greedy generation. In contrast, a large $T$ leads to a uniform averaging and sampling over the whole vocabulary set for CIPHER and natural language generation, respectively. Choosing proper temperatures for the debaters plays a pivotal role in the performance of CIPHER and natural language debate. Thus, to ensure fairness of our empirical evaluation, we utilize Bayesian optimization \citep{bayesopt} to select the best performing temperatures for each method in our experiments in Section \ref{sec:experiments}. Moreover, we conduct sensitivity analysis on the temperatures in Section \ref{sec:temp_ablation}.

 \noindent{\bf Intuition.} The idea behind CIPHER is connected to Expected SARSA \citep{sutton1998reinforcement} in reinforcement learning. Specifically, Expected SARSA replaces the sampled Q-values in vanilla SARSA \citep{rummery1994line,sutton1995generalization} with the expected values over all next actions, leading to significant advantages over vanilla SARSA \citep{van2009theoretical}. For causal LLMs, the autoregressive token generation process can be viewed as a Markov decision process where, at the step $t$, the state is the previously generated response $\response^{(1:t-1)}$, the action space is the vocabulary set $\cV$, the policy is $p^{(t)}$ defined in Eq.~\ref{eq:genprob}, and the reward is tied to the response's accuracy. To this end, our weighted averaging of embeddings shares the same spirit as Expected SARSA, computing expectations over possible tokens. Meanwhile, the natural language response generation process, where tokens are probabilistically sampled, aligns with vanilla SARSA.

% \addtolength{\headsep}{-0.45in}
\section{Experiments}\label{sec:experiments}

\subsection{Experimental Setup}

\noindent{\bf Baseline methods.} We benchmark our proposed approach against the following three baselines:
\begin{itemize}[leftmargin=*]
    \item \textbf{Single Answer}: a single LLM provides one response to the given question in natural language.
    \item  \textbf{Self-Consistency} \citep{wang2022self}: a single LLM independently generates multiple responses to the given question, then applies majority voting to determine the final answer.
    \item \textbf{Natural Language Debate (NLD)}  \citep{du2023improving}: each LLM first provides an initial response to the given question. Subsequently, the LLMs use each other's responses to refine their previous responses (see Appendix \ref{appendix:procedure} for the formal algorithm). Note this approach serves as the most direct baseline for CIPHER (ours), differing primarily in terms of the communication protocol.
\end{itemize}
% \newpage
% \addtolength{\headsep}{0.45in}
 Due to the difficulties experienced by open-source models in generating appropriately formatted feedback as a critic, as discussed in Section \ref{sec:related_work}, we do not include self-improvement via feedback methods (\eg, \citet{madaan2023selfrefine}) in the baselines. For all the methods, we use Bayesian optimization to select temperatures, which are reported in Appendix \ref{appendix:model} for reproducibility of our results.

\noindent{\bf Models.} We conduct most of our experiments using LLaMA2-70B \citep{touvron2023llama}, as it is one of the largest open-source models with an extended context window of up to 4,096 tokens. Additionally, to test our approach's robustness and generalizability, we conduct experiments with various other models, including LLaMA-65B \citep{touvron2023llama1}, Falcon-40B-Instruct \citep{penedo2023refinedweb}, MPT-30B \citep{MosaicML2023Introducing}, and WizardMath-70B-V1.0 \citep{luo2023wizardmath,xu2023wizardlm}.

\noindent{\bf Datasets.} We evaluate CIPHER Debate on five reasoning datasets that span across four different domains. (i) \textbf{GSM8K} \citep{cobbe2021training} consists of a variety of grade school math problems created by human problem writers. 
% Our evaluation is conducted on 200 questions that are randomly sampled from a pool of 1,319 test questions, while \citet{du2023improving} only evaluated 100 of those questions. 
(ii) MMLU \citep{hendrycks2020measuring} we pick three datasets from three different categories, \textbf{Formal Logic} dataset from the Humanities category, \textbf{High School Math} dataset from the STEM category, and \textbf{Professional Psychology} dataset from the Social Science category. (iii) \textbf{Arithmetic}: following \citet{du2023improving}, we evaluate mathematical expressions comprising six unique two-digit numbers that include addition, multiplication, and subtraction operations. For large datasets (GSM8K, Professional Psychology, and Arithmetic), we tune the temperature on a validation set of 200 sampled questions and evaluate on another 200 questions in a separate test set.

 \noindent{\bf Prompts for initial and debate rounds.} We combine few-shot examples with chain-of-thought prompting \citep{wei2022chain} and zero-shot instruction (\enquote{Let's think step by step}) \citep{kojima2022large}  to encourage agents to generate both the final answer and the reasoning steps. This boosts response accuracy and provides valuable information for other LLMs during debates. Additionally, CIPHER is compatible with various prompting methods. See Appendix \ref{appendix:prompt} for detailed prompts.

\noindent{\bf Metrics.} For debates among identical LLMs with different temperatures, we measure the accuracy of the final answer. To evaluate the efficacy of debates among different LLMs that share the same tokenizer, we show more detailed results by testing the correctness of all the final-round answers.

\subsection{Results}
Following \citet{du2023improving}, we assess the performance of CIPHER debates against the baselines in $3$-round debates between $2$ LLMs.  Below we provide a detailed discussion of our results.  

\textbf{Comprehensive evaluation with the LLaMA family.} We comprehensively evaluate CIPHER debate using the LLaMA family of LLMs (LLaMA-65B \citep{touvron2023llama1} and LLaMA2-70B~\citep{touvron2023llama}) across five reasoning datasets. Table \ref{table:llama} presents the results from debates between two identical LLaMA family LLMs operating at different temperatures. Both Self-Consistency (Major@5) and NLD~\citep{du2023improving} exhibit significant performance enhancements over a single answer. Remarkably, CIPHER consistently outperforms both baselines, achieving a $1.0-5.0\%$ boost over NLD across all datasets. We evaluate debates based on the final responses of the agent with a lower temperature, resulting in five responses per debate. For fair comparisons, our self-consistency baselines (labeled as \textit{Major@5}) also use five responses. Additionally, while all the baseline methods display high variance ($0.5 - 3.0\%$ across datasets) due to their token sampling process, CIPHER's deterministic embedding generation ensures consistent outputs.

\setlength{\tabcolsep}{6pt} % Default value: 6pt
\begin{table}[tb]
\vskip-0.45in
\caption{{\bf Debate accuracies (\%) between two identical LLaMA family models at different temperatures with 3 rounds.} Except \textit{Single Answer} baseline, each baseline generates 5 responses per question. Both Self-Consistency (\textit{Major@5}) and \textit{NLD} improve the performance over the \textit{Single Answer} baseline. CIPHER further widens the gap, outperforming NLD by $1.0-5.0$\% consistently.}
\vskip-0.1in
\centering

\resizebox{\textwidth}{!}{

\begin{tabular}{l|lccccc}
\toprule
\multirow{1}{*}{Model} & \multirow{1}{*}{Method} & GSM8K & H.S. Math & Psychology & Formal Logic & Arithmetic \\
\midrule
\multirow{4}{*}{LLaMA2-70B}& Single Answer & 60.0$\pm$\footnotesize{2.3} &  38.3$\pm$\footnotesize{2.6} & 73.6$\pm$\footnotesize{1.2}  & 46.0$\pm$\footnotesize{2.9} & 79.5$\pm$\footnotesize{0.3}   \\ 
 % \hline
 &  Major@5~\citep{wang2022self} & 64.3$\pm$\footnotesize{1.4} & 41.3$\pm$\footnotesize{1.5} & 74.0$\pm$\footnotesize{0.7} & 44.4$\pm$\footnotesize{2.3} & 79.7$\pm$\footnotesize{0.3}   \\ 
 % \hline
  % &  Self-refine   &  -     \\ 
 % \hline
 &  NLD~\citep{du2023improving} & 64.8$\pm$\footnotesize{2.4} & 39.4$\pm$\footnotesize{0.9} & 74.2$\pm$\footnotesize{0.7} & 49.2$\pm$\footnotesize{0.9}  & 81.1$\pm$\footnotesize{0.8}   \\

&   CIPHER ({ours}) & \textbf{66.0}$\pm$\footnotesize{0.0} & \textbf{41.5}$\pm$\footnotesize{0.0} & \textbf{75.0}$\pm$\footnotesize{0.0}  & {\bf 52.4}$\pm$\footnotesize{0.0} & \textbf{85.0}$\pm$\footnotesize{0.0}   \\
\midrule
\multirow{4}{*}{LLaMA-65B} & Single Answer & 50.8$\pm$\footnotesize{1.6} & 33.8$\pm$\footnotesize{1.8} & 68.8$\pm$\footnotesize{1.5}  & 43.5$\pm$\footnotesize{2.7} &  27.6$\pm$\footnotesize{1.1}\\ 
 % \hline
 &  Major@5~\citep{wang2022self} & 52.7$\pm$\footnotesize{3.3} & 36.7$\pm$\footnotesize{0.7} &  70.5$\pm$\footnotesize{0.4} & 46.8$\pm$\footnotesize{2.1} & 29.8$\pm$\footnotesize{0.9}\\ 
 % \hline
 &  NLD ~\citep{du2023improving}  & 51.7$\pm$\footnotesize{1.4} & 36.7$\pm$\footnotesize{0.9} & 70.0$\pm$\footnotesize{2.0} & 46.0$\pm$\footnotesize{1.7}  & 30.4$\pm$\footnotesize{0.4} \\
&  CIPHER ({ours}) & {\bf 52.9}$\pm$\footnotesize{0.0} & {\bf 38.5}$\pm$\footnotesize{0.0} & {\bf 70.9}$\pm$\footnotesize{0.0}  & {\bf 50.8}$\pm$\footnotesize{0.0} &\textbf{33.0}$\pm$\footnotesize{0.0}\\
 \bottomrule
\end{tabular}
}
%\vskip-0.1in
\label{table:llama}
% \vskip-0.2in

% \resizebox{\textwidth}{!}{

% \begin{tabular}{l|lcccccc} 
% \multirow{2}{*}{Model} & \multirow{2}{*}{Method} & GSM8K & High School & Psychology & Formal Logic & Arithmetic & Biographies \\
% & & & Math & & & & \\
% \midrule
% \multirow{4}{*}{LLaMA2-70B}& Single Answer & 59.2 &  38.6 & 71.8  & 45.8 & 76.5  & - \\ 
%  % \hline
%  &  Major@5~\citep{wang2022self} & 65.5 & 41.5 & \textbf{74.0} & 42.9 & 77.8  & - \\ 
%  % \hline
%   % &  Self-refine & - & - &  - & - & -  & - \\ 
%  % \hline
%  &  Natural Language Debate~\citep{du2023improving} & 67.0 & \textbf{44.1} & 73.0 & 49.2  & 83.0  & - \\

% &   CIPHER Debate ({ours}) & \textbf{68.0} & 43.7 & \textbf{74.0}  & {\bf 52.4} & \textbf{86.5}  & - \\
% \midrule
% \multirow{4}{*}{LLaMA-65B} & Single Answer & 50.5 & 33.5 & 66.5  & 43.5 &  29.8 & -\\ 
%  % \hline
%  &  Major@5~\citep{wang2022self} & 57.8 & 36.7 &  67.0 & 44.4 & 31.0 & -\\ 
%  % \hline
%  &  Natural Language Debate~\citep{du2023improving}  & 55.5 & 36.7 & 68.5 & 47.6  & 35.0  & -\\
% &  CIPHER Debate ({ours}) & {\bf 58.8} & {\bf 38.5} & {\bf 70.5}  & {\bf 50.0} &\textbf{ 36.5 }& -\\
%  \bottomrule
% \end{tabular}
% }

\end{table}

Table \ref{tab:llama1_vs_llama2} presents the results of debates between LLaMA-65B and LLaMA2-70B on Arithmetic and GSM8K datasets. While the tokenizer is shared between LLaMA-65B and LLaMA2-70B, they use distinct embeddings for each vocabulary. To tackle this issue, we keep a mapping $\vocab \to [\mathtt{emb_{LLaMA1}}, \mathtt{emb_{LLaMA2}}]$ for each vocabulary and compute the weighted average of the embeddings using the embeddings of the receiver. For example, to pass the message to LLaMA2-70B, we average over $\mathtt{emb_{LLaMA2}}$ during the CIPHER response generation from LLaMA-65B debater. This guarantees that the output of LLaMA-65B is encoded within the LLaMA2-70B's token embedding space.
While debating proves beneficial for both agents, LLaMA-65B experiences a more substantial improvement, especially on Arithmetic dataset, with an increase from 35\% to 62.5\% (Table \ref{tab:llama1_vs_llama2}(a), CIPHER).

\setlength{\tabcolsep}{6pt} % Default value: 6pt
\begin{table}
\caption{\textbf{Debate accuracies (\%) between LLaMA2-70B and LLaMA-65B} on \textbf{(a) Arithmetic} and \textbf{(b) GSM8K} datasets. \textit{Agreement} indicates the consensus (\%) of the two debaters (i.e., they generate the same response). CIPHER debates result in higher accuracies in both debaters compared to NLD.} 

% We report the proportions of the changes in responses compared to the previous round in brackets. 
\vskip-0.1in

\begin{subtable}{1\textwidth}
\sisetup{table-format=-1.2}   % 2 decimals, leave space for minus sign
\centering
\resizebox{\textwidth}{!}{
   \begin{tabular}{lccc|ccc}
\toprule
\multirow{2}{*}{\textbf{(a)} \textbf{Arithmetic}} & \multicolumn{3}{c|}{NLD~\citep{du2023improving}} & \multicolumn{3}{c}{CIPHER (ours)} \\
\cmidrule{2-4} \cmidrule{5-7}
& LLaMA2-70B & LLaMA-65B & Agreement & LLaMA2-70B & LLaMA-65B & Agreement \\
\midrule
Round 1 & 73.0 & 32.5 & 31.5 & 73.5 & 35.0 & 34.5\\
Round 2 & 69.0 & 56.0  & 61.5 & 70.0 & 61.5 & 69.0\\
Round 3 & 72.0 & 60.5 & 77.5 & \textbf{74.5} & \textbf{62.5} & 78.0\\
\bottomrule
\end{tabular}
}
   % \caption{First subtable}
\label{tab:llama1_vs_llama2_arithmetic}
\end{subtable}

\begin{subtable}{1\textwidth}
\sisetup{table-format=4.0} % integer values only, up to 4 digits
\centering
\resizebox{\textwidth}{!}{
   \begin{tabular}{lccc|ccc}
\toprule
\multirow{2}{*}{\textbf{(b)} \textbf{ GSM8K }} \text{   }& \multicolumn{3}{c|}{NLD~\citep{du2023improving}} & \multicolumn{3}{c}{CIPHER (ours)} \\
\cmidrule{2-4} \cmidrule{5-7}
& LLaMA2-70B & LLaMA-65B & Agreement & LLaMA2-70B & LLaMA-65B & Agreement \\
\midrule
Round 1 & 61.0 & 51.3 & 51.5 & 61.5 & 48.0 & 44.0\\
Round 2 & 58.8 & 57.5 & 77.5 & 64.3 & 60.3 & 70.3\\
Round 3 & 61.8 & 59.0 & 88.8 & \textbf{64.8} & \textbf{63.3} & 83.8\\
\bottomrule

\end{tabular}
}
   % \caption{Second subtable}
   \label{tab:llama1_vs_llama2_gsm8k}
\end{subtable}
%\vskip -0.1in

\label{tab:llama1_vs_llama2}
\end{table}

\begin{figure}[t!]
% \vskip-0.1in

  \begin{minipage}[c]{0.47\textwidth}
    \includegraphics[width=\textwidth]{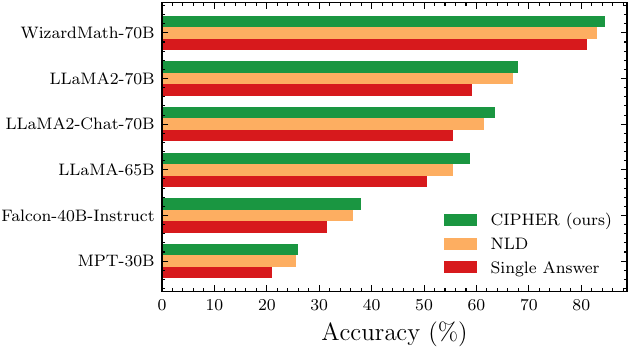}
  \end{minipage}\hfill
  \begin{minipage}[c]{0.5\textwidth}
    \caption{\textbf{Multiagent debates across different models on GSM8K.} Debate in Natural Language (NLD)~\citep{du2023improving} improves the performance of various open-source models compared to the direct generation of a single answer. CIPHER provides an additional performance boost, ranging from 0.5\% to 3.5\% across different models.} \label{fig:more_models}
  \end{minipage}
\vskip -0.15in
\end{figure}

\textbf{Different LLMs on selected datasets.}
To substantiate the efficacy of CIPHER further, we conduct additional debates using three other open-source LLMs: Falcon-40B-Instruct
~\citep{penedo2023refinedweb}, MPT-30B \citep{MosaicML2023Introducing}, and WizardMath-70B-V1.0 \citep{luo2023wizardmath,xu2023wizardlm}. The experiments are performed on GSM8K dataset, as shown in Fig.~\ref{fig:more_models}. NLD~\citep{du2023improving} enhances the performance of a wide range of open-source models compared to generating a single answer. CIPHER further provides additional performance boosts ranging from 0.5\% to 3.5\% across various models, which are significant considering there is no modification to the model weights.

\section{Analysis and Discussion}\label{sec:analysis}

In this section, we provide experimental results of debates in extended scales, debate temperature sensitivity analysis, and an ablation study through partial CIPHER implementation. Additional experimental results on positional bias and performance bounds of debates are in Appendix \ref{appendix:extra_exp}. We also present qualitative analysis of CIPHER debates on a few selected questions in Appendix \ref{appendix:qual}.

\subsection{Debate in Extended Scales}
\begin{figure}[t]
\vskip-0.4in
\centering
\begin{subfigure}[t]{0.4\columnwidth}
    \centering
    \includegraphics[width=\textwidth]{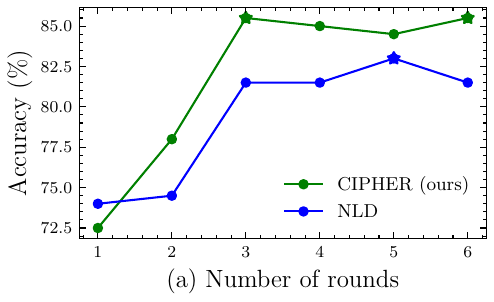}
    \label{fig:num_rounds}
\end{subfigure}
 \qquad 
\begin{subfigure}[t]{0.4\columnwidth}
    \centering
    \includegraphics[width=\textwidth]{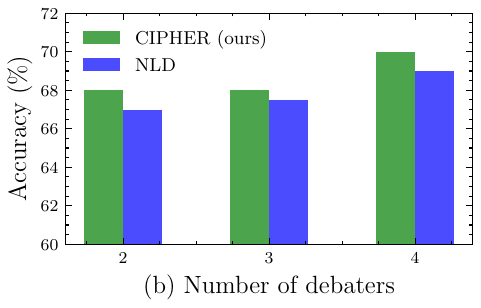}
    \label{fig:num_debaters}
\end{subfigure}
\vskip-0.2in
\caption{\textbf{(a) Number of rounds.} In general, adding more rounds helps to boost performance on both CIPHER and NLD~\citep{du2023improving}. Stars ({\Large$\filledstar$}) indicate the best performance of 2 LLaMA2-70B debaters on Arithmetic dataset. \textbf{(b) Number of debaters.} Adding more debaters is beneficial for both CIPHER and NLD~\citep{du2023improving}. We report the accuracies of LLaMA2-70B on GSM8K dataset.}
\label{fig:num_rounds_and_debaters}
\end{figure}

While recent work has shown the phenomenon of rapidly diminishing marginal return of scaling up the numbers of debaters and rounds~\citep{du2023improving,liang2023encouraging}, we nonetheless include these scenarios for the sake of a comprehensive evaluation. This allows us to paint a more complete picture of the performance landscape of CIPHER debate. Fig.~\ref{fig:num_rounds_and_debaters}a shows that adding more debate rounds can eke out a small gain for debate methods at the cost of more responses. Likewise, from Fig.~\ref{fig:num_rounds_and_debaters}b, we observe that involving more debaters is beneficial to the debate with a similar trade-off.

\subsection{Temperature Sensitivity}\label{sec:temp_ablation}
\begin{figure}[t]
\vskip -0.4in
\centering
\begin{subfigure}[t]{0.32\columnwidth}
    \centering
    \includegraphics[width=\textwidth]{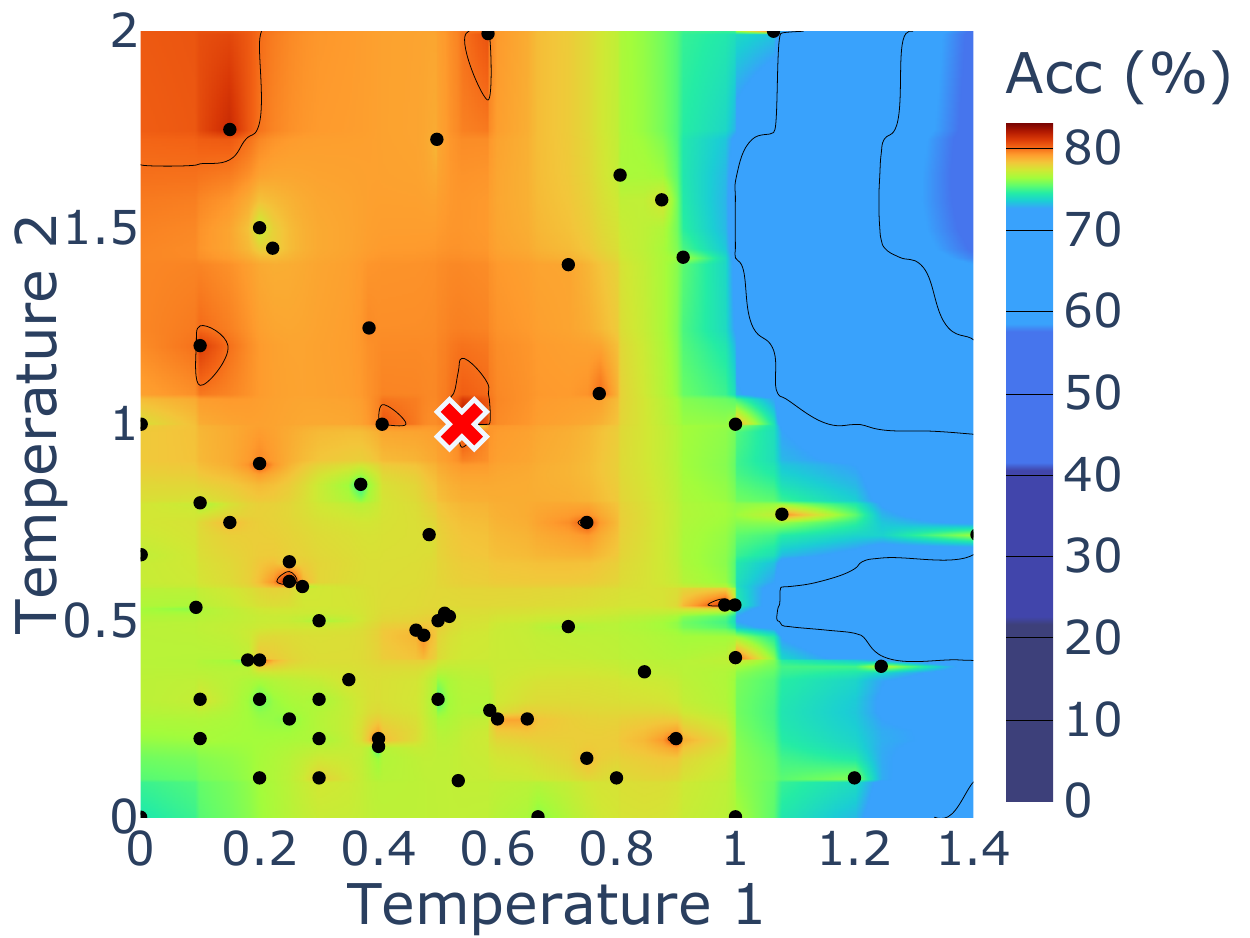}
    % \caption{Arithmetic}
    \label{fig:arithmetic_human}
\end{subfigure}
\begin{subfigure}[t]{0.32\columnwidth}
    \centering
    \includegraphics[width=\textwidth]{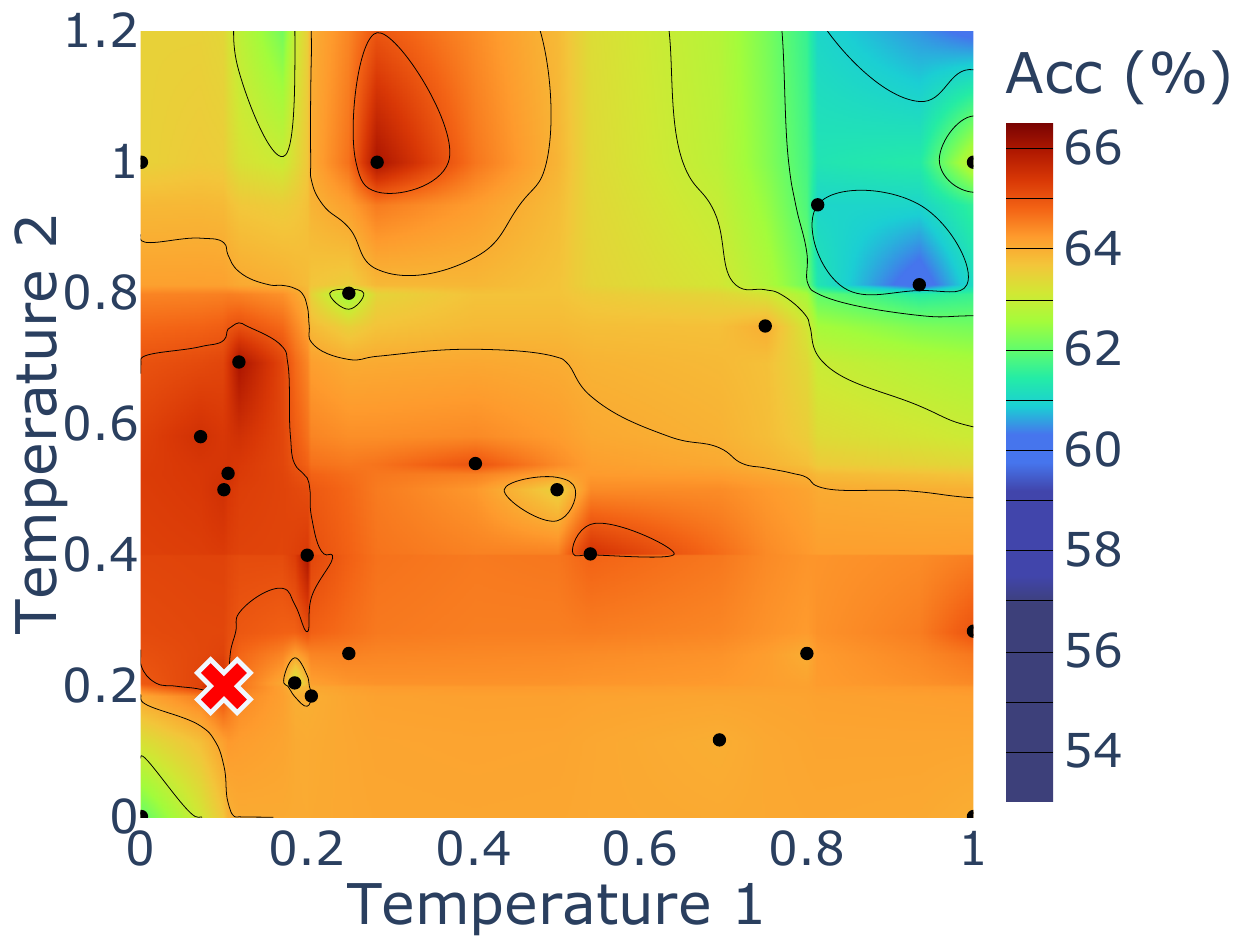}
    % \caption{GSM8K}
    \label{fig:gsm8k_human}
\end{subfigure}
\begin{subfigure}[t]{0.32\columnwidth}
    \centering
    \includegraphics[width=\textwidth]{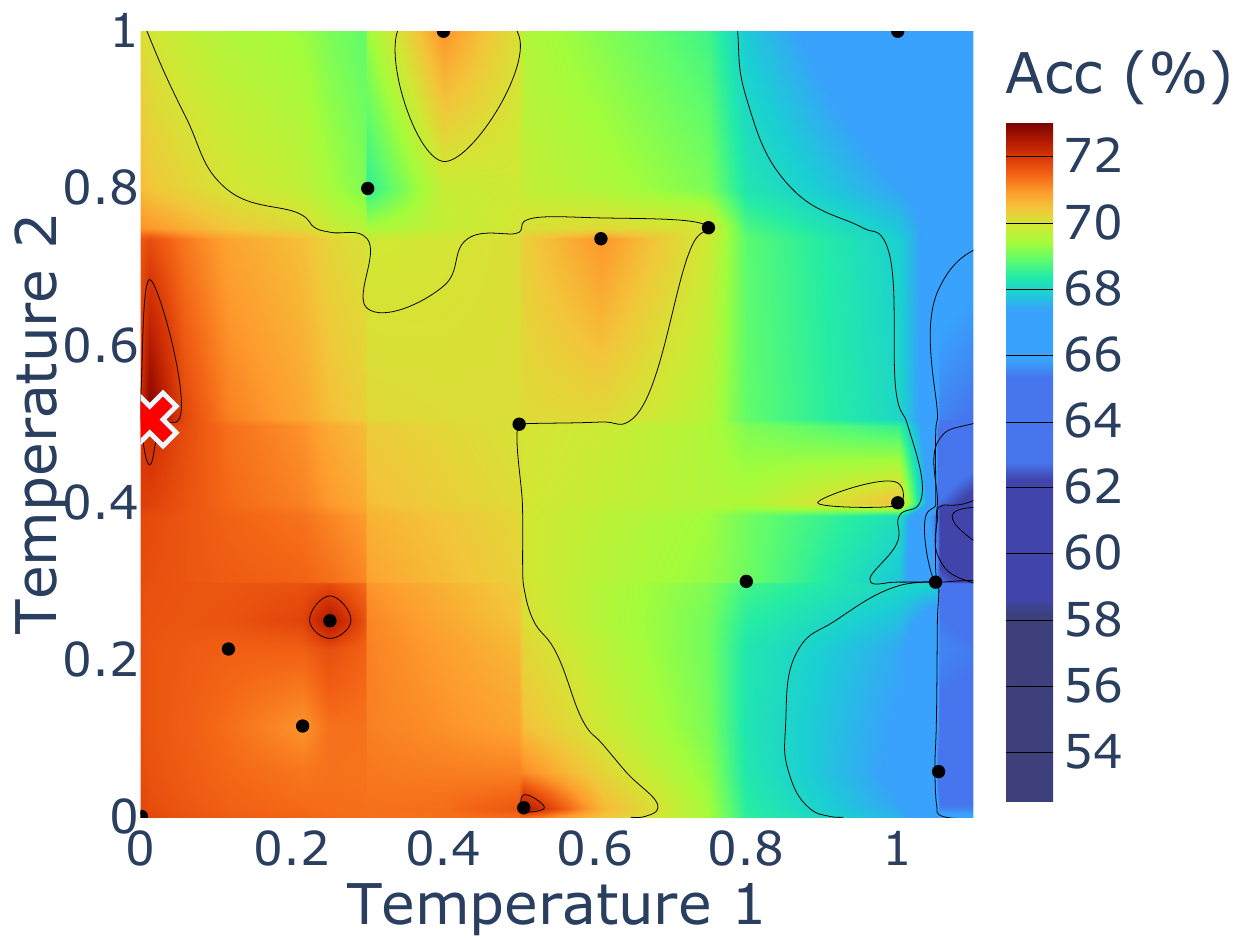}
    % \caption{Professional Psychology}
    \label{fig:mmlu_math_human}
\end{subfigure}
\begin{subfigure}[t]{0.32\columnwidth}
    \centering
    \includegraphics[width=\textwidth]{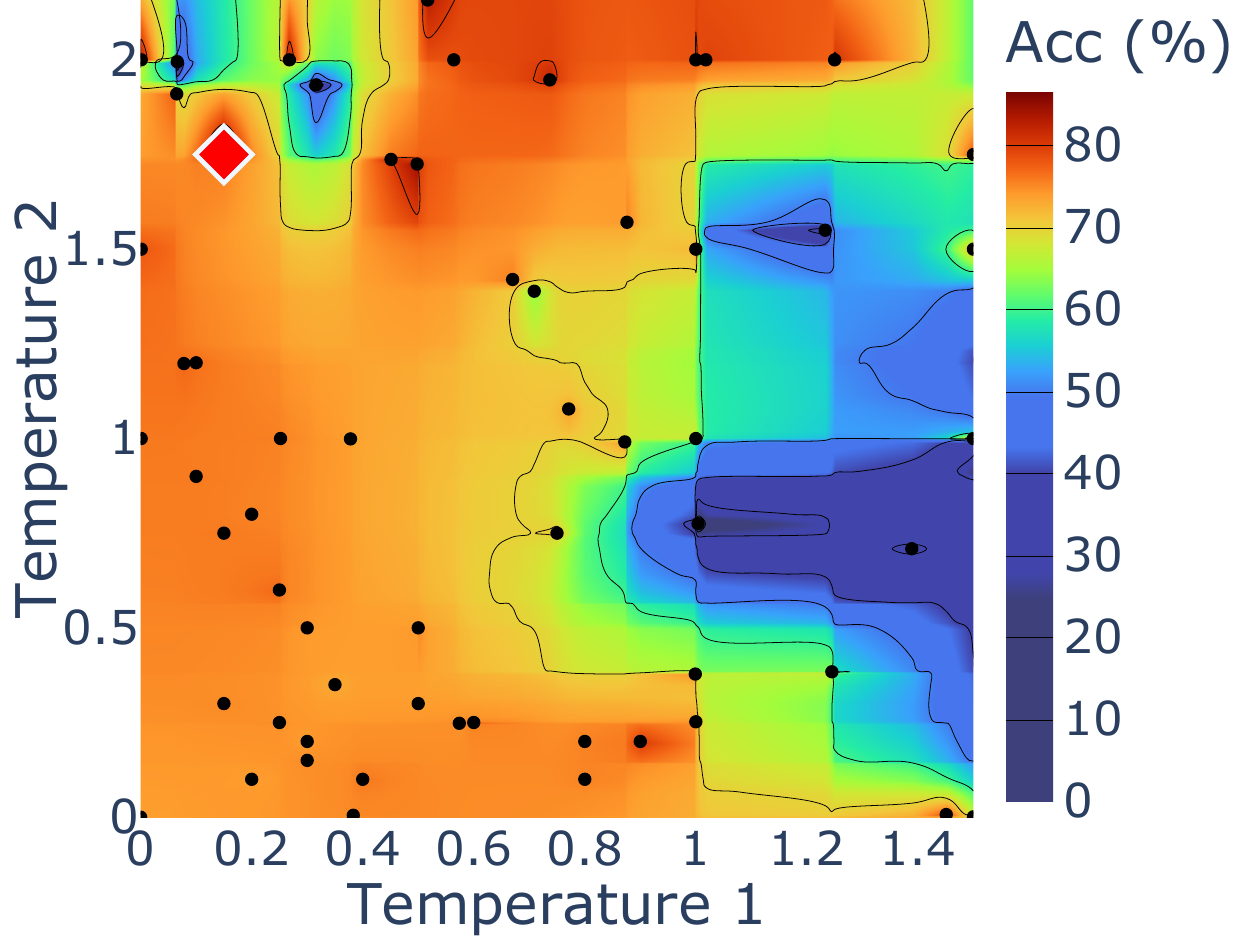}
    \caption{Arithmetic}
    \label{fig:arithmetic}
\end{subfigure}
\begin{subfigure}[t]{0.32\columnwidth}
    \centering
    \includegraphics[width=\textwidth]{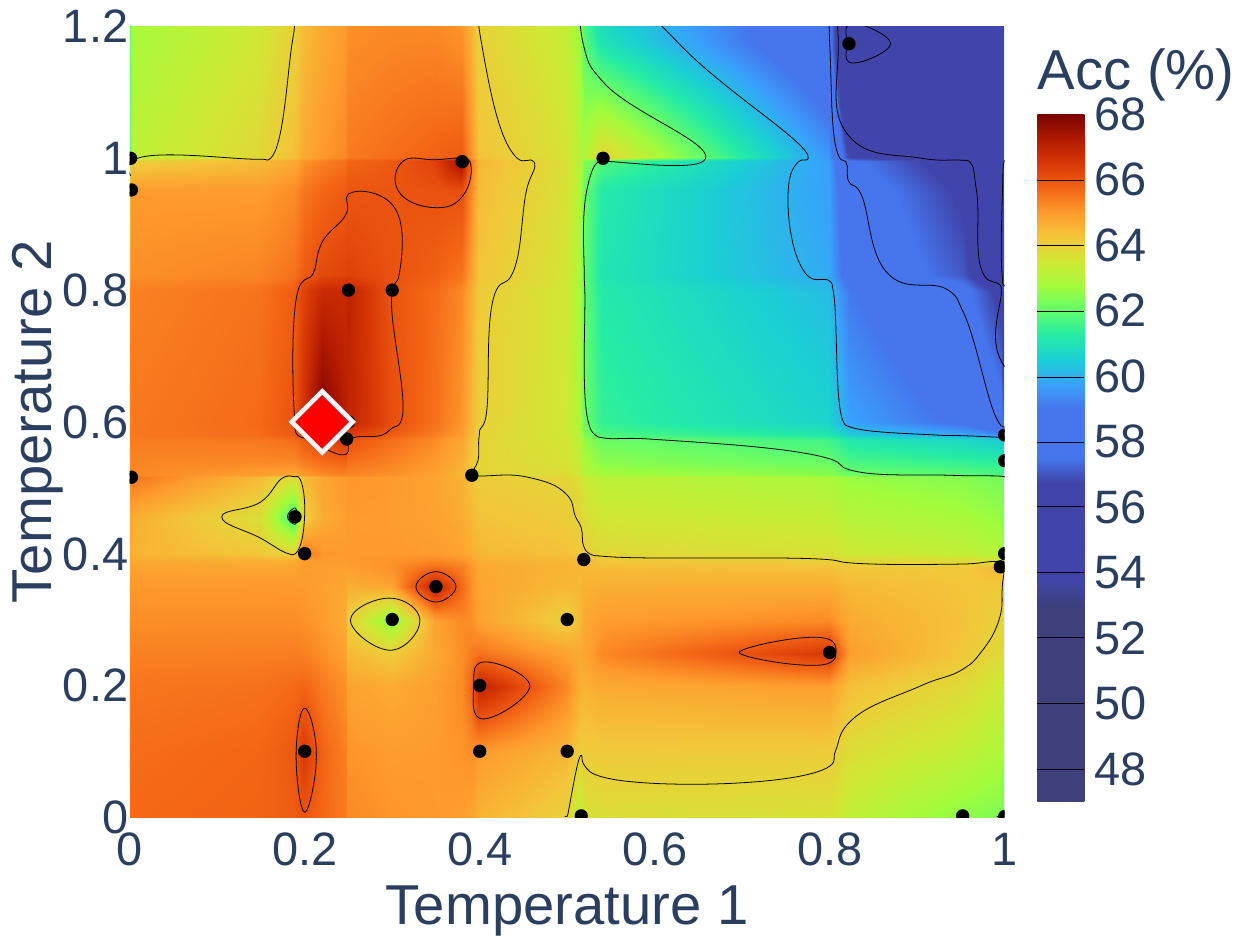}
    \caption{GSM8K}
    \label{fig:gsm8k}
\end{subfigure}
\begin{subfigure}[t]{0.32\columnwidth}
    \centering
    \includegraphics[width=\textwidth]{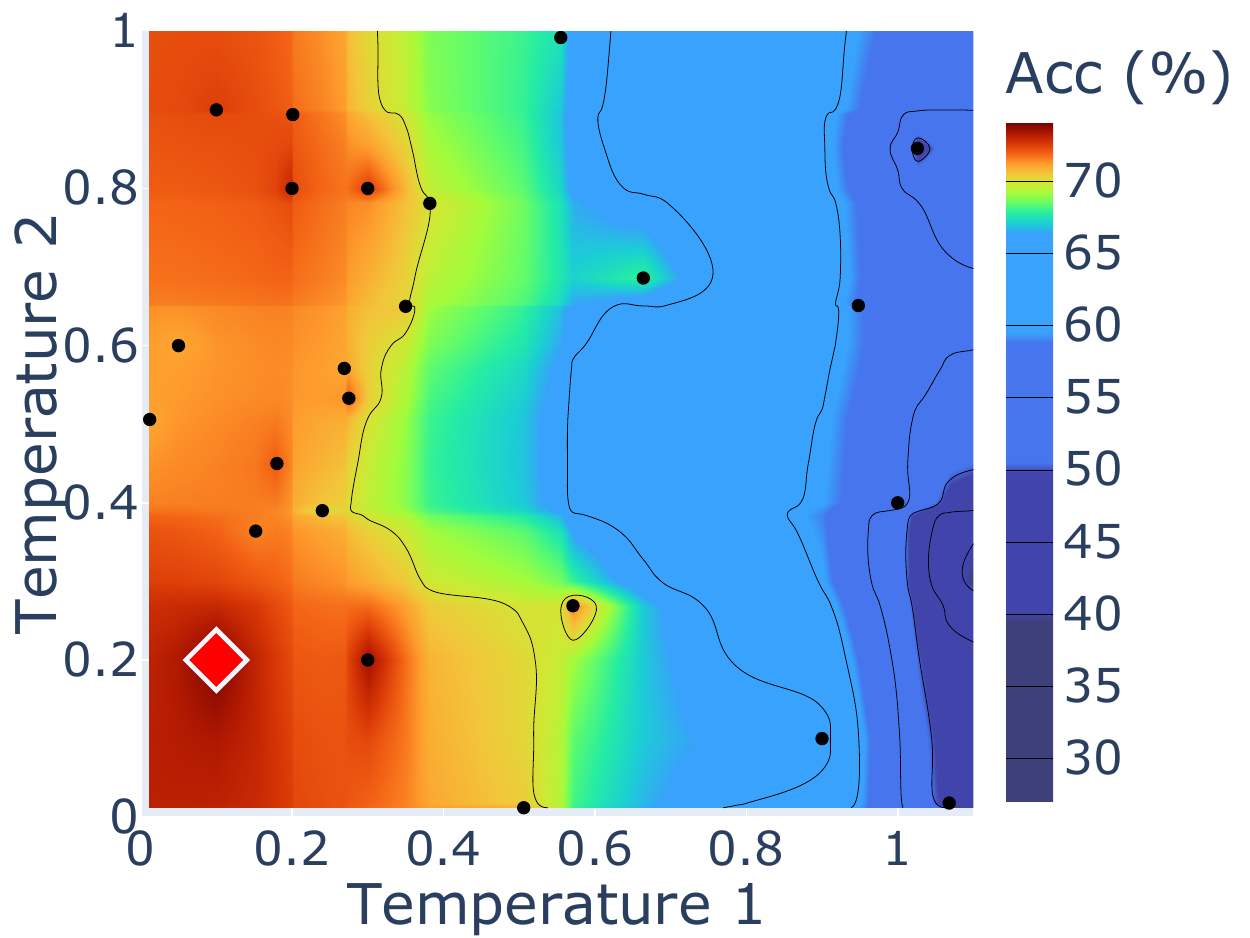}
    \caption{Professional Psychology}
    \label{fig:mmlu_math}
\end{subfigure}
\vskip-0.05in
\caption{\textbf{2D-contour plots of accuracy over different temperatures.} We depict the performance debates between two LLaMA2-70B CIPHER debaters across different pairs of temperatures, where cross marks (\textcolor{red}{\xmark}) and diamonds (\reddiamond) denote the best performance of NLD~\citep{du2023improving} (top row) and CIPHER (ours, bottom row), respectively. We report the debate performance based on the final responses generated by debater 1 (\textit{temperature 1)}. 
% We can observe that the best performance is achieved when \textit{temperature 1} is lower than \textit{temperature 2}.
For CIPHER (bottom row), the optimal regions appear on the left side of the charts, where \textit{temperature 1} is lower than \textit{temperature 2}. 
These indicate that a good strategy is using various temperature agents, and choosing the response of the lower temperature agent as the final debate answer. See Section \ref{sec:temp_ablation} for more discussion.}
\label{fig:contour_plots}
% \vskip-0.05in
\end{figure}

In this section, we delve into the performance dynamics of debates regarding temperature selection (Fig.~\ref{fig:contour_plots}).
In particular, we investigate the advantages of allowing certain debaters to deviate from natural language during debates. More concretely, one debater operates at a lower temperature to ensure the final answer remains comprehensible to humans, while the other debater is tasked with conveying information that may deviate from natural language by generating its responses at a higher temperature. 
We employ Bayesian optimization \citep{bayesopt} to identify promising pairs of temperatures in debates between two LLaMA2-70B debaters on Arithmetic dataset. Unlike the final response aggregation strategy in CIPHER, we evaluate the effectiveness of these temperature pairs based on the accuracy of the final response from the first debater (\textit{temperature 1}), which can operate at a higher temperature. Such an experiment setting not only sheds light on the impact of temperature selection on debate performance but also guides our final response aggregation strategy.

In general, the debate is more beneficial when the debaters are more diverse.
% i.e., the difference in their temperatures is large. 
For NLD~\citep{du2023improving} (Fig.~\ref{fig:contour_plots}, top row), optimal performance is often achieved with all the temperatures set below $1$. This can be attributed to the inherent information loss during the token sampling process in natural language generation. At higher temperatures, LLMs are more likely to produce nonsensical responses, potentially weakening stronger debaters' performance.
In contrast, for CIPHER, optimal performance is often obtained at wider spread apart temperatures, as shown in Fig.~\ref{fig:contour_plots}a, bottom row. Notably, the optimal regions for CIPHER are mostly on the left side of the charts, indicating that CIPHER benefits the most when it pairs a low-temperature agent with a high-temperature one. At higher temperatures, the probability distribution of the tokens in the vocabulary becomes more uniform, allowing CIPHER's responses to lean towards less confident token choices. This effectively complements the information communicated by the other lower-temperature debater, which focuses on more confident tokens. Additionally, low-temperature agents are necessary for collecting results that can be interpretable by humans. Therefore, a good strategy for CIPHER is to employ various temperature agents and use the response of the lower temperature agent as the final debate answer.

\begin{figure}[tb]
% \vskip-0.1in
  \begin{minipage}[c]{0.41\textwidth}
    \caption{\textbf{Partial CIPHER.} We invoke CIPHER for token generation in some steps while opting for greedy sampling in others, based on generation uncertainty. Experiments with LLAMA2-70B pairs on the Arithmetic dataset.} \label{fig:partial_cipher}
  \end{minipage}
\hfill
  \begin{minipage}[c]{0.53\textwidth}
    \includegraphics[width=\textwidth]{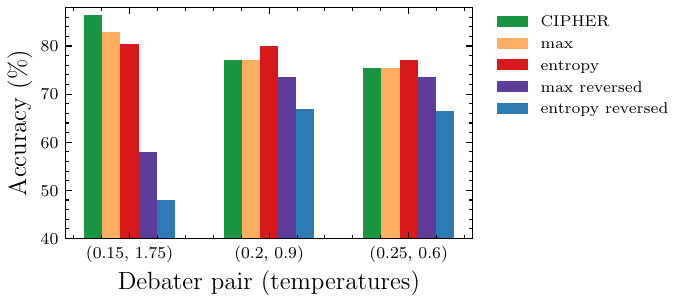}
  \end{minipage}
% \vskip-0.05in
\end{figure}

\subsection{Ablation study on CIPHER}\label{sec:partial_cipher_sec}
To unveil the mechanisms contributing to the performance gain brought by CIPHER, we invoke CIPHER response generation only at the positions where the model exhibits high uncertainty $U^{(t)}$ regarding the next token. Setting a threshold $\varepsilon > 0$, we adopt CIPHER generation whenever $U^{(t)} > \varepsilon$ and default to greedy sampling otherwise. To quantify the uncertainty $U^{(t)}$ at a given position $t$, we employ two variants (i) \textit{entropy} $U^{(t)} = -\sum_{i = 1}^V p^{(t)}_i \log p^{(t)}_i$ of the generation distribution $p^{(t)}$ over the vocabulary set, or (ii) \textit{max} probability $U^{(t)} = 1 - \max_{i \in [V]} p^{(t)}_i$. Additionally, we explore two reversed variants, \textit{entropy reversed} and \textit{max reversed}, which invoke CIPHER when $U^{(t)} \leq \varepsilon$. 
% Such targeted applications of CIPHER serve to shed light on the importance of information retention during moments of model uncertainty. 
Fig.~\ref{fig:partial_cipher} shows that partial application of CIPHER in \textit{max} and \textit{entropy} aligned closely with full CIPHER application, endorsing our hypothesis that CIPHER advantages from the information retention during the moments of uncertainty. In contrast, \textit{entropy reversed} and \textit{max reversed} see huge performance drops, showing that diverging from CIPHER can drastically diminish the efficacy of debates.

\section{Conclusion}\label{sec:conclusion}
Our proposed CIPHER demonstrates promising results across various reasoning tasks. Without necessitating special training, CIPHER enhances the debate efficacy of a wide range of LLMs, outperforming majority voting and debate methods using natural language. Our study highlights the potential of LLMs' belief information and serves as the first step towards unlocking this potential in LLM debates, closing the gap between proprietary models and open-source models. It remains intriguing whether there is an even more efficient way of transferring belief information among LLMs, given the fact that LLMs are only trained to intake the embeddings of natural language text tokens.

\newpage
\section*{Limitations and Broader Impacts}
One limitation of CIPHER lies in its applicability, which is currently restricted to Language Models (LLMs) sharing a common vocabulary set. Expanding our approach to encompass LLMs with distinct tokenizers would require meticulous alignment of their respective embedding-vocabulary mappings. Such alignment is challenging, as tokenizers often employ significantly different text segmentation strategies. For instance, while one tokenizer may break a word into subwords or characters, another might treat it as a single token. We recognize this as a promising avenue for future research. Overcoming this limitation could pave the way for constructing even more robust and efficient LLM agent systems, potentially unlocking unprecedented collaborative capabilities among diverse LLMs. On a positive note, the open-source LLM community is seeing rapid growth of families of open-source LLMs with shared tokenizers such as LLaMA family (LLaMA, LLaMA2, LLaMA2-chat, and Code LLaMA). As such, we are optimistic that our work, in its current form, will continue to contribute to the growth and evolution of the community.

%\newpage

\bibliography{multiagent_debate}
\bibliographystyle{iclr2024_conference}

\appendix
\newpage
\section{Multiagent Debate Framework}\label{appendix:procedure}

We demonstrate the procedure of natural language debate in Algorithm \ref{alg_debate}. We use $\|$ to denote concatenation. Here, the debaters $\{D_i\}_{i \in [n]}$ are conventional LLMs and the responses $\response_i$ are in natural language. In contrast, the debaters $\{D_i\}_{i \in [n]}$ in Algorithm \ref{alg_cipher} are CIPHER debaters and the responses $\emb\response_i$ are embeddings generated by \eqref{eq:emb-response}. We also note that, since the responses generated in Algorithm \ref{alg_debate} are all in natural language, we can directly aggregate (denoted as \texttt{Aggregate}) the final round responses without performing a nearest neighbor search over the vocabulary set.

\begin{algorithm}[H]
\caption{Multiagent Natural Language Debate}
\label{alg_debate}
\begin{algorithmic}
{\small
\STATE {\bf Input:}
Question and instructions $\prompt$, number of rounds $R \geq 2$, and $n$ LLM debaters $\{D_i\}_{i \in [n]}$.

{\bf For debater $i=1,2,\dots,n$:}

{\addtolength{\leftskip}{0.2in}

\STATE Get initial natural language response $\response_i \leftarrow D_i(\prompt)$ from debater $i$.

}

{\bf EndFor}

% Debate rounds
{\bf For round $r=1,\dots,R$:}
 
{\addtolength{\leftskip}{0.2in}

\STATE Get updated prompt $\prompt \leftarrow \mathrm{concat}(\prompt, \response_1,  \dots, \response_n)$.

}

{\addtolength{\leftskip}{0.2in}

{\bf For debater $i=1,\dots,n$:}

}

{\addtolength{\leftskip}{0.4in}

\STATE Get updated natural language response $\response_i \leftarrow D_i(\prompt)$.

}

{\addtolength{\leftskip}{0.2in}

{\bf EndFor}

}

{\bf EndFor}

\STATE  $\response^\star \leftarrow \texttt{Aggregate}(\response_1 , \dots , \response_n) $

\STATE {\bf Output:} Final response $\response^\star$
}

\end{algorithmic}
\end{algorithm}

\section{Extra Experiments}\label{appendix:extra_exp}

\subsection{Investigating Positional Bias in Debates}
\begin{figure}[H]
\begin{center}
\centerline{\includegraphics[width=0.5\columnwidth]{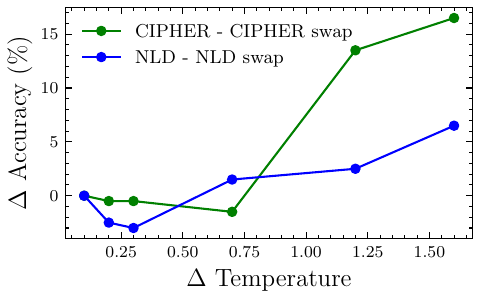}}
\caption{\textbf{Effect of the order of responses in two-agent debate.} The green line shows the difference between CIPHER and its variant when swapping the order of responses (CIPHER swap), while the blue line shows the difference between NLD~\citep{du2023improving} and its variant (NLD swap).}
\label{fig:pos_bias}
\end{center}
\end{figure}

The issue of positional bias in utilizing LLMs as evaluators has attracted increasing attention in recent studies \citep{wang2023large,zheng2023judging,liusie2023zero}. Although our multiagent debate setting differs from ``LLMs as evaluators,'' we recognize that the sequence in which prior rounds responses are fed into subsequent rounds of debates could still have non-negligible effects on the outcomes of the debates. Recall that in debate rounds, for CIPHER, we feed the responses of other debaters first, then the response of the debater itself.  In \Cref{fig:pos_bias}, we provide a further investigation into the positional bias within our multiagent debate setting by swapping the order, \ie, a variant of CIPHER where other debaters' responses are fed after its own response. We find that the effect of positional bias is negligible when the two debaters operate at similar temperatures. However, when the debaters are dissimilar, swapping the order of responses can result in a significant difference. Both NLD~\cite{du2023improving} and CIPHER show better performance when the debaters are more diverse.

\subsection{Performance Bounds of Debates}
\begin{figure}[H]
\begin{center}
\centerline{\includegraphics[width=1.0\columnwidth]{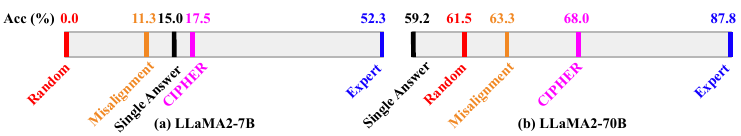}}
\caption{\textbf{Upper bound and lower bound of two-agent CIPHER debate on GSM8K.} A debater generates with \textcolor{redcolor}{\textbf{Random}} (Extremely high temperature debater), \textcolor{orangecolor}{\textbf{Misalignment}} (debater responds misaligned ground-truth), \textcolor{black}{\textbf{Single Answer}}, \textcolor{violetcolor}{\textbf{CIPHER}} (ours), and \textcolor{blue}{\textbf{Expert}} (debater responds ground-truth) on GSM8K dataset. We observe that debate can be detrimental when the model has low capacity (Fig. \textit{a}), but it does not pose much harm in the case of a more powerful model (Fig. \textit{b}).}
\label{fig:upper_lower_bounds}
\end{center}
\end{figure}

Prior work showcases significant performance enhancements achieved through LLM debates. Yet, the limit of such a performance gain remains a captivating facet of study. To investigate the performance upper bound that can be achieved through multiagent debate, we conduct an experiment where the LLM debater is pitted against an expert debater. We proxy the expert debater by having it always give the ground truth answers. Conversely, to establish the performance lower bound, we conduct an experiment where the LLM debater constantly receives nonsensical feedback from other debaters. Specifically, we employ 2 dummy debaters, one with extremely high temperature that gives non-sense, and the other gives non-relevant responses by using misaligned ground truth answers from other questions in the batch. 
Figure \ref{fig:upper_lower_bounds} illustrates these bounds on GSM8K with LLaMA2-70B and its much less capable version, LLaMA2-7B.

\subsection{Attention Heatmaps}
\begin{figure}[H]
\begin{center}
\centerline{\includegraphics[width=1.0\columnwidth]{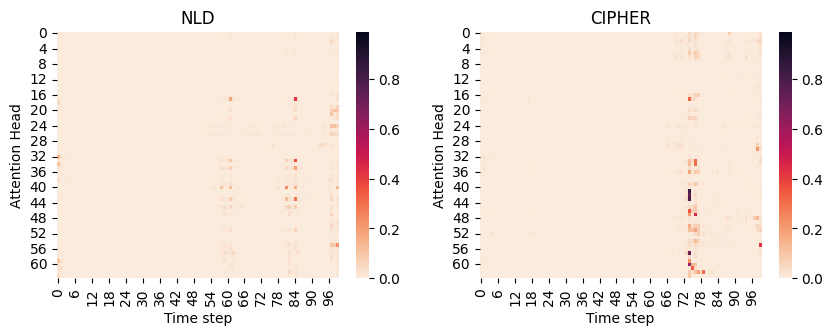}}
\caption{\textbf{Attention heatmaps of CIPHER and NLD~\citep{du2023improving} when using 2 LLaMA2-70B.} The heatmap generated by CIPHER indicates some significant bright spots, especially around the 40$^{th}$ attention head and the 74$^{th}$ time steps.}
\label{fig:attention_heatmap}
\end{center}
\end{figure}

Figure \ref{fig:attention_heatmap} shows a comparison of attention heatmaps for CIPHER and NLD~\citep{du2023improving} at the 45$^{th}$ decoder layer of LLaMA2-70B. These heatmaps correspond to the arithmetic question we used in Figure~\ref{fig:debate_demo} during the last debate round. Specifically, we compute the similarity between the $q$ vector of the last token of the first agent and the $k$ vectors of its preceding 100 tokens. We can observe that NLD's heatmap exhibits uniform attention distribution, lacking intense focus on any particular segment. Conversely, CIPHER's heatmap shows some distinct bright spots, particularly around the 40$^{th}$ attention head and the 74$^{th}$ time step. This suggests that the model's attention is highly focused on those areas, potentially indicating areas of higher relevance for the task.

\section{Qualitative Results}\label{appendix:qual}
\begin{figure}[H]
\begin{center}
\centerline{\includegraphics[ width=0.8\columnwidth]{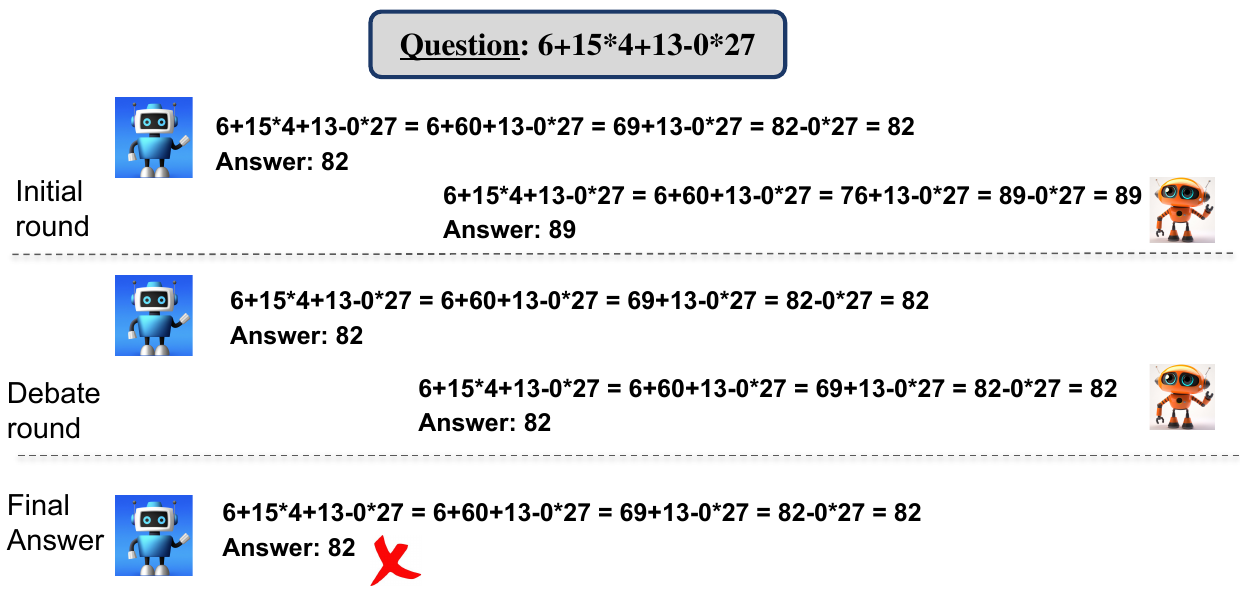}}
% \vskip-0.1in
\caption{\textbf{Example of NLD on Arithmetic dataset.} A complete debate involving two LLaMA2-70B debaters using a temperature pair of (0.00, 0.67), as presented in \Cref{fig:debate_demo}.
}
\label{fig:example_human_arith1}
\end{center}
\end{figure}

\begin{figure}[H]
\begin{center}
\centerline{\includegraphics[width=0.8\columnwidth]{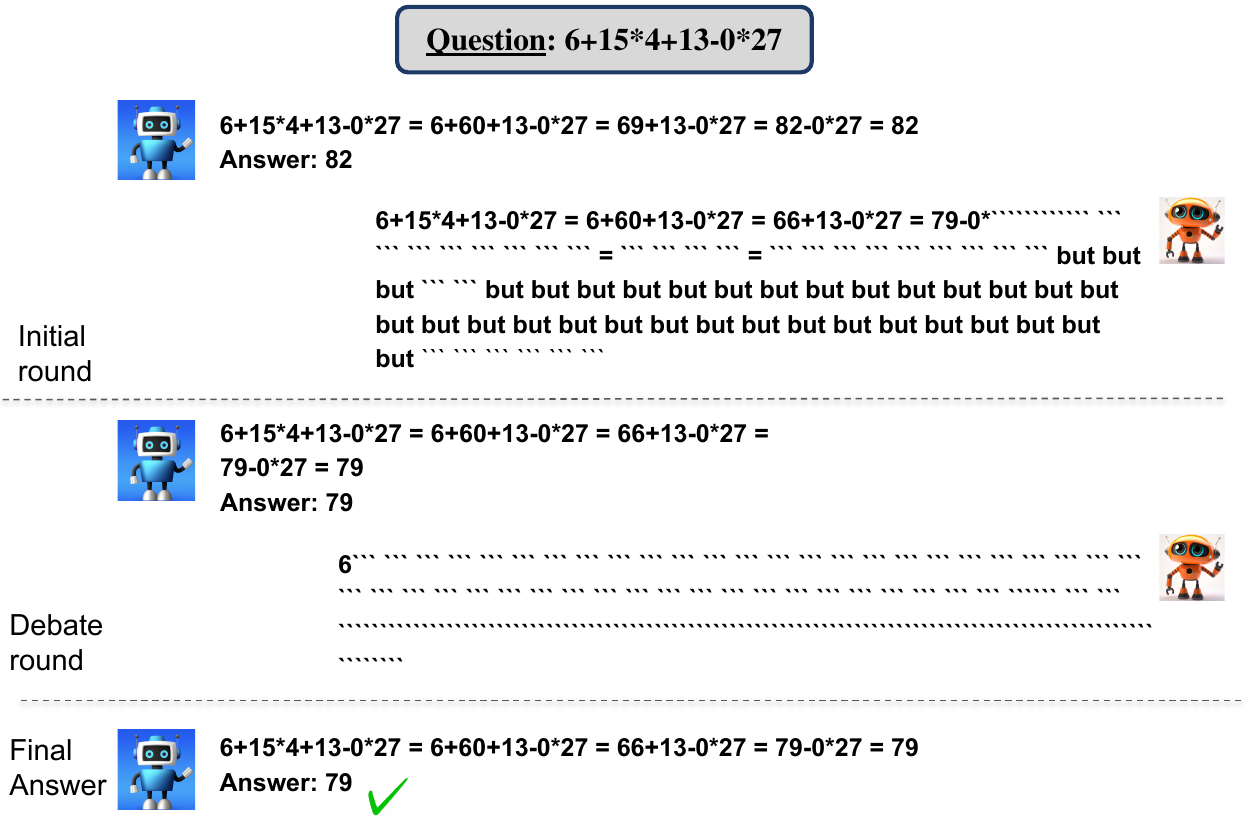}}
% \vskip-0.1in
\caption{\textbf{Example of CIPHER on Arithmetic dataset.}  A complete debate involving two LLaMA2-70B debaters using a temperature pair of (0.25, 1.75), as presented in \Cref{fig:debate_demo}. We convert the generated embeddings back to natural language using a nearest neighbor search on the vocabulary.
}
\label{fig:example_vector_arith1}
\end{center}
\end{figure}

\begin{figure}[H]
\begin{center}
\centerline{\includegraphics[width=1\columnwidth]{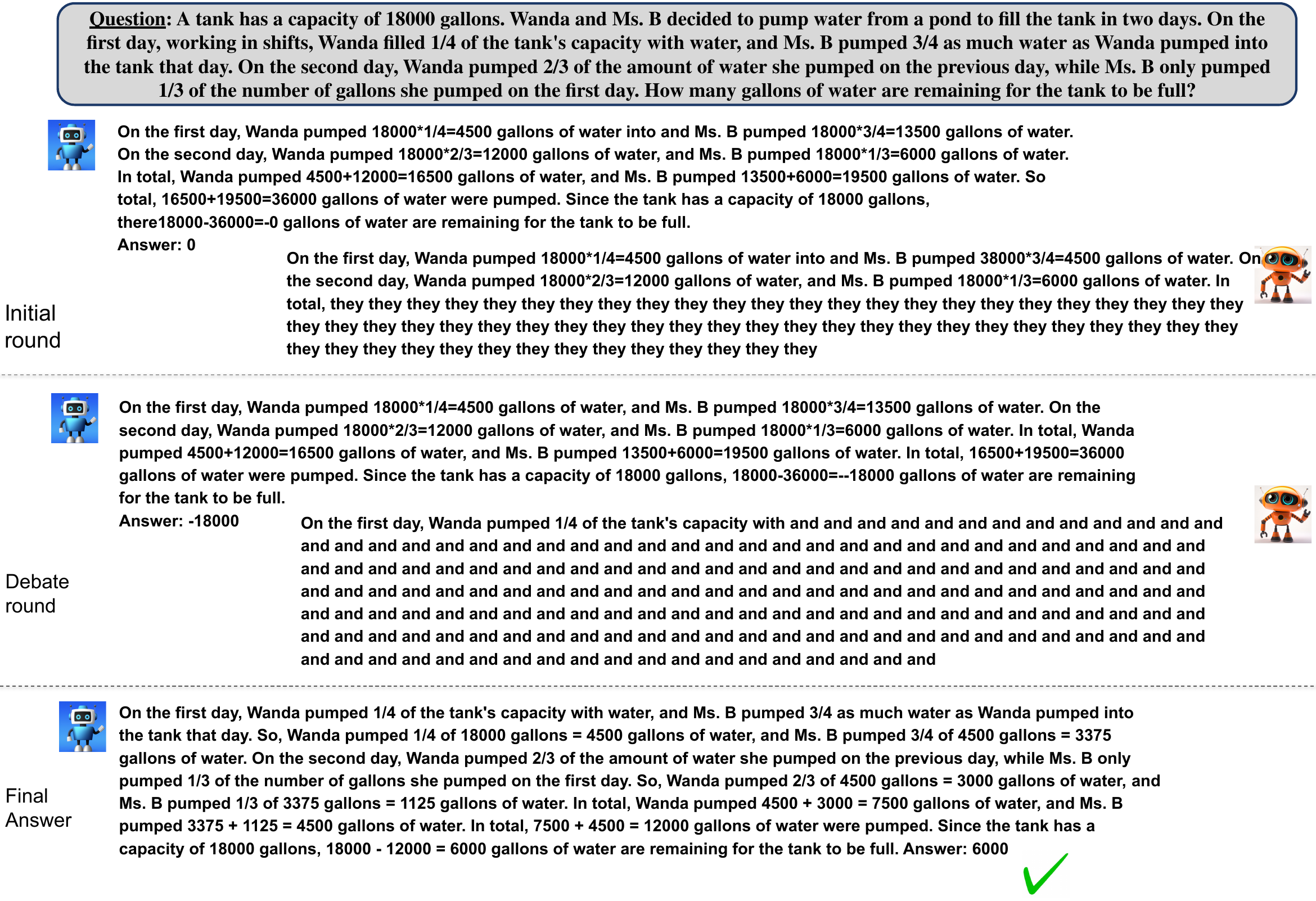}}
% \vskip-0.1in
\caption{\textbf{Example of CIPHER on GSM8K dataset.}  A complete debate involving two LLaMA2-70B debaters using a temperature pair of (0.40, 1.00).
}
\label{fig:gsm8k_vec_example}
\end{center}
\end{figure}

We present some detailed results of each debater's response within the context of a two-agent debate of NLD~\citep{du2023improving} and our method, CIPHER. 
\Cref{fig:example_human_arith1} and
\Cref{fig:example_vector_arith1} display
the complete example used to demonstrate the content shown in \Cref{fig:debate_demo} in the main paper.
In \Cref{fig:gsm8k_vec_example}, we present a debate utilizing CIPHER on the GSM8K dataset.

\section{Detailed Experiment Setups}\label{appendix:model}

\subsection{Temperature}
\label{appendix:temperature}
To ensure the reproducibility of our results, we include the temperatures of the debaters in each of our experiments in the following.

{\bf Temperatures in the comprehensive evaluation with LLaMA family LLMs. }Corresponding to Table \ref{table:llama}, Table \ref{table:llama_temp} presents the temperatures of LLaMA family debaters during the debates.

\setlength{\tabcolsep}{6pt} % Default value: 6pt
\begin{table}[H]

\centering
\caption{Temperatures of two identical LLaMA family models during debates.}
\label{table:llama_temp}

\resizebox{\textwidth}{!}{
\begin{tabular}{l|lccccc} 
\toprule
Model & Method  & GSM8K & High School Math & Psychology & Formal Logic & Arithmetic \\
\midrule
\multirow{4}{*}{LLaMA2-70B}& Single Answer & 0.15 &  0.24 & 0.33  & 0.30 & 0.35 \\
 % \hline
 &  Major@5~\cite{wang2022self} & 0.80 & 0.30 & 0.30 & 0.50 & 0.60 \\ 

 &  NLD~\cite{du2023improving} &(0.10, 0.20) & (0.30, 0.49) & (0.01, 0.51) & (0.10, 0.20)  & (0.54, 1.00) \\
&   CIPHER ({ours}) &   (0.22, 0.60) & (0.10, 0.82) & (0.10, 0.20)  & (0.10, 0.20) & (0.15, 1.75) \\
\midrule
\multirow{4}{*}{LLaMA-65B}& Single Answer & 0.20 &  0.30 & 0.10  & 0.20 & 0.40 \\ 
 % \hline
 &  Major@5~\cite{wang2022self} & 0.80 & 0.80  & 0.80 & 0.60 & 0.80 \\ 

 &  NLD~\cite{du2023improving} & (0.10, 0.20) & (0.40, 0.50) & (0.30, 0.40) & (0.20, 0.85)  & (0.08, 0.20) \\

&   CIPHER ({ours}) & (0.25, 0.85) & (0.49, 0.80) & (0.10, 0.40) & (0.25, 0.85) & (0.67, 1.43) \\
 \bottomrule
\end{tabular}
}

\end{table}

Corresponding to Table \ref{tab:llama1_vs_llama2}, Table \ref{tab:llama1_vs_llama2_temp} presents the temperatures of LLaMA-65B and LLaMA2-70B during the debates on the Arithmetic and GSM8K datasets.

\setlength{\tabcolsep}{6pt} % Default value: 6pt
\begin{table}[H]
\caption{Temperatures of LLaMA2-70B and LLaMA-65B during the dabtes on \textbf{(a) Arithmetic} and \textbf{(b) GSM8K} datasets.} 
\centering
\begin{subtable}{0.75\textwidth}
\sisetup{table-format=-1.2}   % 2 decimals, leave space for minus sign
\centering
\resizebox{\textwidth}{!}{
   \begin{tabular}{lcc|cc}
\toprule
\multirow{2}{*}{\textbf{(a)} \textbf{Arithmetic}} & \multicolumn{2}{c|}{ NLD~\cite{du2023improving}} & \multicolumn{2}{c}{CIPHER (ours)} \\
\cmidrule{2-3} \cmidrule{4-5}
& LLaMA2-70B & LLaMA-65B & LLaMA2-70B & LLaMA-65B \\
\midrule
Temperature & 0.32 & 0.51 & 0.35 & 0.35 \\
\bottomrule
\end{tabular}
}
   % \caption{First subtable}
\label{tab:llama1_vs_llama2_arithmetic_temp}
\end{subtable}

\bigskip
\begin{subtable}{0.75\textwidth}
\sisetup{table-format=4.0} % integer values only, up to 4 digits
\centering
\resizebox{\textwidth}{!}{
   \begin{tabular}{lcc|cc}
\toprule
\multirow{2}{*}{\textbf{(b)} \textbf{ GSM8K }} \text{   }& \multicolumn{2}{c|}{ NLD~\cite{du2023improving}} & \multicolumn{2}{c}{CIPHER (ours)} \\
\cmidrule{2-3} \cmidrule{4-5}
& LLaMA2-70B & LLaMA-65B & LLaMA2-70B & LLaMA-65B \\
\midrule
Temperature & 0.20 & 0.40 & 0.35 & 0.63 \\
\bottomrule

\end{tabular}
}
   % \caption{Second subtable}
   \label{tab:llama1_vs_llama2_gsm8k_temp}
\end{subtable}

\label{tab:llama1_vs_llama2_temp}
\end{table}

\noindent{\bf Temperatures in evaluation across different LLMs.} Corresponding to Figure \ref{fig:more_models}, Table \ref{table:all_model_temp} presents the temperatures of open-source LLMs on GSM8K dataset.

\setlength{\tabcolsep}{6pt} % Default value: 6pt
\begin{table}[H]

\centering
\caption{Temperatures of 2-agent debate across different models on GSM8K}
\label{table:all_model_temp}

\resizebox{\textwidth}{!}{

\begin{tabular}{l|cccccc} 
\toprule
Method  & LLaMA2-70B & LLaMA2-Chat-70B & LLaMA-65B & Falcon-40B-Instruct & MPT-30B & WizardMath-70B \\
\midrule
Single Answer & 0.15 &  0.15 & 0.20  & 0.40 & 0.45  & 0.00 \\ 
 % \hline
NLD~\citep{du2023improving} &  (0.10, 0.20)  &   (0.20, 0.40) &  (0.10, 0.20)  & (0.20, 0.40) & (0.35, 0.62)  & (0.15, 0.35) \\ 

CIPHER (ours) & (0.22, 0.60) &  (0.25,0.65) & (0.25, 0.85)  & (0.25, 0.65) & (0.23, 0.64) & (0.26, 0.69) \\ 
% \midrule
% \multirow{3}{*}{LLaMA-65B} & Single Answer & 50.5 & 33.5 & 66.5  & 43.5 &  29.8 & -\\ 
%  % \hline
%  &  Major@5 & 57.8 & 36.7 &  67.0 & 44.4 & 31.0 & -\\ 
%  % \hline
%  &  Natural Language Debate  & 55.5 & 36.7 & 68.5 & 47.6  & 35.0  & -\\
% &  CIPHER Debate (\textbf{ours}) & {\bf 58.8} & {\bf 38.5} & {\bf 70.5}  & {\bf 50.0} &\textbf{ 36.5 }& -\\
 \bottomrule
\end{tabular}
}

\end{table}

\noindent{\bf Temperatures in debates in extended scales. } Corresponding to Figure \ref{fig:num_rounds_and_debaters}, we report the temperatures of LLaMA2-70B debaters in the debates in extended scales in Table \ref{table:extended_scale_temp_num_rounds} and Table \ref{table:extended_scale_temp_num_debaters}.

\setlength{\tabcolsep}{6pt} % Default value: 6pt
\begin{table}[H]
\centering

\caption{Temperatures of LLaMA2-70B debaters in debates in extended scales on GSM8K dataset in reported \Cref{fig:num_rounds_and_debaters}(a).}
\label{table:extended_scale_temp_num_rounds}

\resizebox{0.38\textwidth}{!}{
\begin{tabular}{l|ccc} 
\toprule
Method  & 2 debaters\\
\midrule
NLD~\citep{du2023improving} &  (0.013, 1.072)   \\ 
CIPHER & (0.498, 1.725)\\ 
 \bottomrule
\end{tabular}
}

\end{table}
\setlength{\tabcolsep}{6pt} % Default value: 6pt
\begin{table}[H]
\centering

\caption{Temperatures of LLaMA2-70B debaters in debates in extended scales on GSM8K dataset reported in \Cref{fig:num_rounds_and_debaters}(b).}
\label{table:extended_scale_temp_num_debaters}

\resizebox{0.8\textwidth}{!}{
\begin{tabular}{l|ccc} 
\toprule
Method  & 2 debaters & 3 debaters & 4 debaters\\
\midrule
NLD~\citep{du2023improving} &  (0.100, 0.500)  &   (0.300, 0.500, 0.700) &  (0.442, 0.176, 0.745, 0.539) \\ 
CIPHER & (0.250, 0.600) &  (0.001, 0.725, 1.067) & (0.641, 0.464, 0.507, 0.202) \\ 
 \bottomrule
\end{tabular}
}

\end{table}

\subsection{Computation Resource}
For LLaMA family debates, we use $4\times$ NVIDIA A100 SXM 80GB GPUs as the major computation resource.

\section{debate prompts}\label{appendix:prompt}
In this section, we provide the detailed prompts we use for each dataset. The prompts consist of two parts:
\begin{itemize}
    \item Initial round prompt: it prompts the debaters to generate their initial response with a chain-of-thought explanation.
    \item debate round prompt: it prompts debaters to incorporate the responses generated in the previous rounds and provide refined responses.
\end{itemize}
%%% GSM8K
\subsection{GSM8K}

\subsubsection{Initial round prompt}

We employ the following 3-shot prompt for all models, with the exception of WizardMath-70B-V1.0 \citep{luo2023wizardmath,xu2023wizardlm}, for which we use the CoT prompt provided by the authors \footnote{\url{https://github.com/nlpxucan/WizardLM/tree/main/WizardMath}}.

\textbf{LLaMA, Falcon, MPT}
\begin{myprompt}
{prompts/gsm8k/init_question_3shot_v3.txt}
\end{myprompt}

\textbf{WizardMath}
\begin{myprompt}
{prompts/gsm8k/init_wizardmath.txt}
\end{myprompt}

\subsubsection{debate round prompt}
We utilize the same 3-shot prompt below for all models, except for WizardMath-70B-V1.0, where we use a zero-shot prompt according to the authors' instructions.

\textbf{LLaMA, Falcon, MPT}
\begin{myprompt}
{prompts/gsm8k/debate_2debaters_v1.txt}
\end{myprompt}

\textbf{WizardMath}
\begin{myprompt}
{prompts/gsm8k/debate_2debaters_wizardmath.txt}
\end{myprompt}

%%% MMLU - Math
\subsection{MMLU - High School Math}

\subsubsection{Initial round prompt}
\begin{myprompt}
{prompts/mmlu/init_high_school_mathematics_v1.txt}
\end{myprompt}

\subsubsection{debate round prompt}
\begin{myprompt}
{prompts/mmlu/debate_high_school_mathematics_2debaters_v1.txt}
\end{myprompt}

%%% MMLU - Psychology
\subsection{MMLU - Professional Psychology}

\subsubsection{Initial round prompt}
\begin{myprompt}
{prompts/mmlu/init_professional_psychology_v1.txt}
\end{myprompt}

\subsubsection{debate round prompt}
\begin{myprompt}
{prompts/mmlu/debate_professional_psychology_2debaters_v1.txt}
\end{myprompt}

%%% MMLU - Formal Logic
\subsection{MMLU - Formal Logic}

\subsubsection{Initial round prompt}
\begin{myprompt}
{prompts/mmlu/init_fomal_logic_v2.txt}
\end{myprompt}

\subsubsection{debate round prompt}
\begin{myprompt}
{prompts/mmlu/debate_fomal_logic_2debaters_v2.txt}
\end{myprompt}

%%% Arithmetic
\subsection{Arithmetic}

\subsubsection{Initial round prompt}
\begin{myprompt}
{prompts/arithmetic/init_prompt.txt}
\end{myprompt}

\subsubsection{debate round prompt}
\begin{myprompt}
{prompts/arithmetic/debate_2debaters_v1.txt}
\end{myprompt}

% % %%%%%%%%%%%%%%%%%%%%%%%%%%%%%%%%%%%%%%%%%%%%%%%%%%%%%%%%%%%%%%%%%%%%%%%%%%%%%%%
% %%%%%%%%%%%%%%%%%%%%%%%%%%%%%%%%%%%%%%%%%%%%%%%%%%%%%%%%%%%%%%%%%%%%%%%%%%%%%%%

\end{document}